\documentclass{article} %
\usepackage{colm2024_conference}

\usepackage{microtype}
\usepackage{hyperref}
\usepackage{url}
\usepackage{booktabs}
\definecolor{darkblue}{rgb}{0, 0, 0.5}
\hypersetup{colorlinks=true, citecolor=darkblue, linkcolor=darkblue, urlcolor=darkblue}

\usepackage{graphicx}
\usepackage{subcaption}
\usepackage{booktabs}
\usepackage{supertabular}
\usepackage{array}
\usepackage{multirow}
\usepackage{wrapfig}
\usepackage{xspace}
\usepackage{adjustbox}

\usepackage{amsthm}

\theoremstyle{definition}
\newtheorem{definition}{Definition}

\newcommand\secref[1]{\S\ref{#1}}

\newcommand\term[1]{\textbf{#1}}
\newcommand\numattr[2]{\texttt{#1:#2}}

\newcommand\model[1]{#1\xspace}
\newcommand\llamasmall{\model{Llama 2 7B}}
\newcommand\llamamed{\model{Llama 2 13B}}
\newcommand\falconsmall{\model{Falcon 7B}}
\newcommand\mistralsmall{\model{Mistral 7B}}

\newcommand{\smallsubfig}[2]{\subcaptionbox{#1}{\includegraphics[width=0.325\textwidth]{#2}}}
\newcommand{\medsubfig}[2]{\subcaptionbox{#1}{\includegraphics[width=0.495\textwidth]{#2}}}

\newcommand\plssubfig[3]{\smallsubfig{#1}{dimreduction#2#3}}

\newcommand\plssubfigs[1]{
	\centering
	\plssubfig{Birthyear}{P569}{#1}
	\plssubfig{Death year}{P570}{#1}
	\plssubfig{Population}{P1082}{#1}
	\plssubfig{Elevation}{P2044}{#1}
	\plssubfig{Latitude}{P625.lat}{#1}
	\plssubfig{Longitude}{P625.long}{#1}
}

\newcommand\dimredsubfig[4]{\smallsubfig{#1}{pls_2d.#2.#3.#4.2.pdf}}
\newcommand\dimredsubfigs[1]{
	\centering
	\dimredsubfig{Birthyear}{P569}{#1}{Oranges}
	\dimredsubfig{Death year}{P570}{#1}{Reds}
	\dimredsubfig{Population}{P1082}{#1}{Greens}
	\dimredsubfig{Elevation}{P2044}{#1}{Greys}
	\dimredsubfig{Latitude}{P625.lat}{#1}{Blues}
	\dimredsubfig{Longitude}{P625.long}{#1}{Purples}
}

\newcommand\editsubfig[3]{\smallsubfig{#1}{edit_curve_#3_#2.pdf}}

\newcommand\editsubfigs[1]{
	\centering
	\editsubfig{Birthyear}{P569}{#1}
	\editsubfig{Death year}{P570}{#1}
	\editsubfig{Population}{P1082}{#1}
	\editsubfig{Elevation}{P2044}{#1}
	\editsubfig{Latitude}{P625.lat}{#1}
	\editsubfig{Longitude}{P625.long}{#1}
}

\newcolumntype{C}[1]{>{\centering\let\newline\\\arraybackslash\hspace{0pt}}m{#1}}

\newcommand*{\tran}{^{\mkern-1.5mu\mathsf{T}}}
\newcommand{\twodots}{\mathinner {\ldotp \ldotp}}

\makeatletter
\DeclareDocumentEnvironment{appendixfig}{}{
   \par\medskip\noindent
   \begin{minipage}{\linewidth}
   \def\@captype{figure}
   \centering
}
{
\end{minipage}
\par\bigskip
}
\makeatother 

\makeatletter
\DeclareDocumentEnvironment{appendixtbl}{}{
   \par\medskip\noindent
   \begin{minipage}{\linewidth}
   \def\@captype{table}
   \centering
}
{
\end{minipage}
\par\bigskip
}
\makeatother 

\title{Monotonic Representation of Numeric Properties in \\ Language Models}

\author{%
Benjamin Heinzerling\textsuperscript{\textnormal{1, 2}} \textnormal{and} Kentaro Inui\textsuperscript{\textnormal{2, 1}} \\
\textsuperscript{\textnormal{1}}RIKEN \& \textsuperscript{\textnormal{2}}Tohoku University \\
\hypersetup{urlcolor=black} \href{mailto:benjamin.heinzerling@riken.jp}{\tt benjamin.heinzerling@riken.jp} \hspace{0.06em} $\vert$ \hspace{0.12em} \href{mailto:inui@tohoku.ac.jp}{\tt inui@tohoku.ac.jp}%
}

\author{%
Benjamin Heinzerling \\
RIKEN / Tohoku University \\
\texttt{benjamin.heinzerling@riken.jp} \\
\AND
Kentaro Inui \\
MBZUAI / Tohoku University / RIKEN\\
\texttt{kentaro.inui@mbzuai.ac.ae} \\
}

\colmfinalcopy

\begin{document}
\maketitle
\begin{abstract}
Language models (LMs) can express factual knowledge involving numeric properties such as \emph{Karl Popper was born in 1902}.
However, how this information is encoded in the model's internal representations is not understood well.
Here, we introduce a simple method for finding and editing representations of numeric properties such as an entity's birth year.
Empirically, we find low-dimensional subspaces that encode numeric properties monotonically, in an interpretable and editable fashion.
When editing representations along directions in these subspaces, LM output changes accordingly.
For example, by patching activations along a "birthyear" direction we can make the LM express an increasingly late birthyear: \emph{Karl Popper was born in 1929}, \emph{Karl Popper was born in 1957}, \emph{Karl Popper was born in 1968}.
	Property-encoding directions exist across several numeric properties in all models under consideration, suggesting the possibility that monotonic representation of numeric properties consistently emerges during LM pretraining. Code: \url{https://github.com/bheinzerling/numeric-property-repr}
\end{abstract}

\section{Introduction: Do LMs represent numeric properties ``appropriately''?}

Language models (LMs) are capable of expressing factual knowledge \citep{petroni2019language,jiang2020know,roberts2020much,heinzerling2021language,kassner2021multilingual}.
For example, when queried \emph{In which year was Karl Popper born?} \model{Llama 2} \citep{touvron2023llama} gives the correct answer \emph{1902}.
While the question if and to what degree LMs can be said to ``know'' anything at all is subject of ongoing debate \citep{bender2020climbing,hase2023beliefs,mollo2023vector,lederman2024language}, empirical work has progressed from behavioral analysis focused on the accuracy and robustness of knowledge expression \citep{shin2020autoprompt,jiang2021know,zhong2021factual,youssef2023facts} to representational analysis aimed at understanding how factual knowledge is encoded\footnote{We say ``X is encoded in Y'' as shorthand for ``X can be easily extracted from Y''. See caveats in \secref{sec:limitations}.} in model parameters \citep{decao2021editing,mitchell2021fast,meng2022locating} and activations \citep{hernandez2023remedi,merullo2023mechanism,geva2023dissecting,gurnee2023language}.

However, representational analysis has so far mainly targeted entity-entity relations such as \emph{Warsaw is the capital of Poland} \citep{merullo2023mechanism}.
If and how LM representations encode factual knowledge involving numeric properties, such as an entity's birthyear, is less understood.
Unlike entity-entity relational knowledge, numeric properties have natural ordering and monotonic structure, such as earlier/later and smaller/larger scales or geographic coordinate systems.
While this kind of structure is natural and intuitive for humans, LMs encounter numeric properties only in form of largely unordered and unstructured textual mentions.
This raises the question if LMs learn to represent numeric properties appropriately, i.e., according to their natural structure.

\begin{figure}[t!]
	\includegraphics[width=\textwidth]{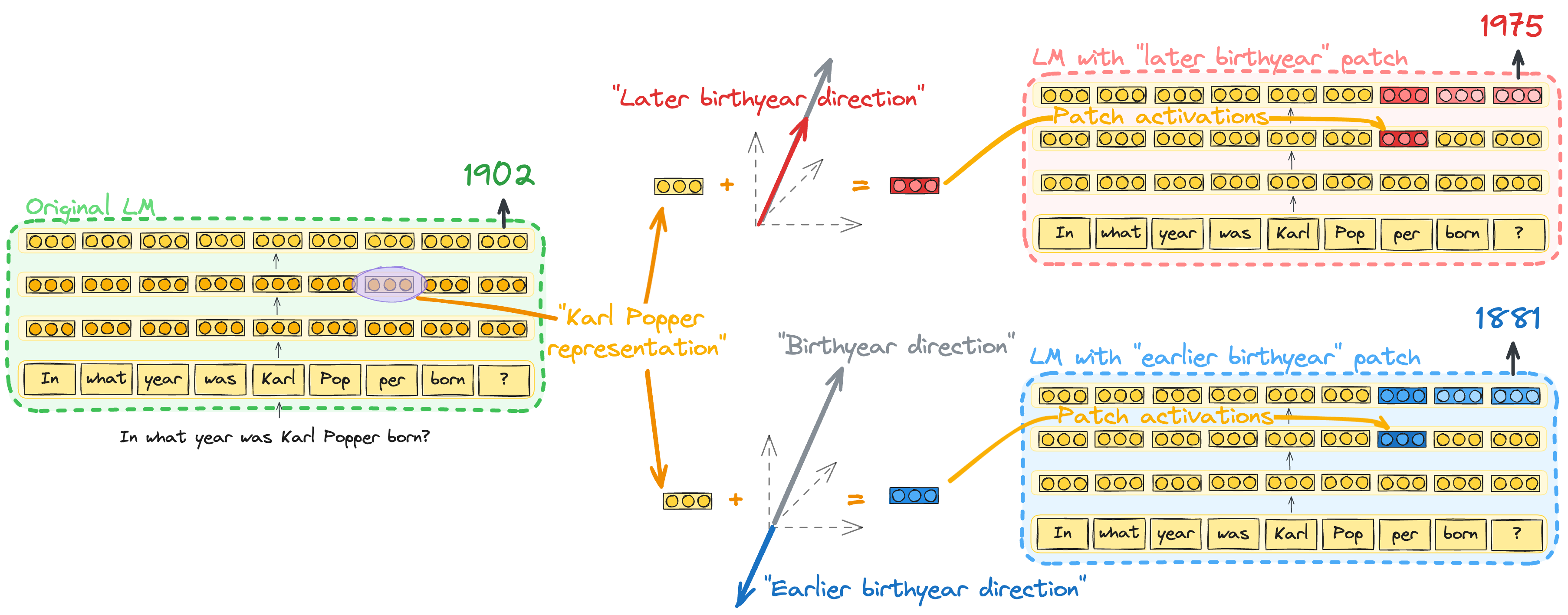}
	\caption{Sketch of our main finding. Patching entity representations along specific directions in activation space yields corresponding changes in model output.}
	\label{fig:overview}
\end{figure}

Here, we devise a simple method for identifying and manipulating representations of numeric properties in LMs.
We find low-dimensional subspaces that strongly correlate with numeric properties across models and numeric properties, thereby confirming and extending prior observations of representations of numeric properties in LMs \citep{lietard2021language,faisal2023geographic,gurnee2023language,godey2024scaling}.
Going beyond prior work (see \secref{sec:relwork}), we show that by causally intervening along certain directions in these subspaces, LM output changes correspondingly.
That is, we find a monotonic relationship between the intervention and the quantity expressed by the LM.
For example, an entitity's year of birth shifts according to the strength and sign of the intervention along a ``birthyear'' direction (Figure~\ref{fig:overview}).
Taken together, our findings suggest that LMs represent numeric properties in a way that reflects their natural structure and that such monotonic representations consistently emerge during LM pretraining.

\paragraph{Terminology.} Before moving to the main part, we briefly clarify important terms.
A \term{quantity} consists of a scalar numeric \term{value} paired with a \term{unit} of measurement.
A \term{numeric property} is a property that can naturally be described by a quantity, e.g., birthyear, population size, geographic latitude.
A \term{numeric attribute} is an instance of a numeric property, associated with a particular entity. For example, Karl Popper has the numeric attribute \numattr{birthyear}{1902}.
By \term{linear representation} we denote the idea that a numeric attribute is encoded in a linear subspace of a LM's activation space.
Finally, a \term{monotonic representation} is a linear representation characterized by a monotonic relationship between directions in activation space and the value of the encoded numeric attribute.
That is, as activations shift along a particular direction the value of the corresponding numeric attribute increases or decreases monotonically.

\section{Finding Property-Encoding Directions}
\label{sec:correlation}

\subsection{Motivation: Representation learning, linear representation hypothesis}
While numeric properties generally can be mapped naturally onto simple canonical structures, such as number lines or coordinate systems, it is not immediately obvious that pretraining on largely unstructured data enables LMs to appropriately represent such structures.
Our main goal is to find out if and how numeric properties, such as an entity's birthyear, are encoded in the geometry of LM representations.
How could such an encoding look like? Based on two arguments, we hypothesize that numeric properties are encoded in low-dimensional linear subspaces of activation space.

The first argument rests on a key principle in representation learning: a model generalizes if and only if its representations reflect the structure of the data \citep{conant1970every,liu2022grokking}.
To the degree that current LMs generalize, in the sense of achieving non-trivial performance on benchmarks involving knowledge of numeric properties \citep{petroni2019language}, we can expect that their representations reflect the structure of numeric properties.
And since the natural structures of many numeric properties are low-dimensional linear spaces such as number lines, we expect to find low-dimensional linear structure in the representations of a well-performing model.

As second argument we adduce the linear representation hypothesis, which posits a correspondence between concepts and linear subspaces \citep{elhage2022toy,park2023linear,nanda2023emergent}.
If the linear representation hypothesis is true,\footnote{For positive evidence, see \citet{marks2023geometry,merullo2023mechanism,tigges2023linear,jiang2024origins}, i.a.} this would imply that numeric properties are encoded in linear subspaces.
For brevity, we will call a low-dimensional linear subspace of a LM's activation space a \emph{direction}, regardless of whether it is one- or multi-dimensional.

\subsection{Method: Partial least squares regression on entity representations}
Motivated by the hypothesis that numeric properties are encoded as directions in activation space, we now devise an experimental setup for finding out if such directions exist.
An obvious choice for identifying linear structure would be principal component analysis \citep[PCA;][]{pearson1901lines}.
However, PCA looks for directions of maximum variance and is unsupervised in the sense that we cannot cannot provide model outputs to steer the algorithm towards particular directions.
What we would like to do instead is to provide supervision in order to find directions in activation space that maximally covary with model outputs.
A standard method applicable to this kind of problem is partial least squares regression.

Partial least squares regression \citep[PLS;][]{wold2001pls} is a supervised method for finding relationships between two data matrices $X$ and $Y$.
Briefly summarizing the detailed exposition by \citet{wold2001pls}, the objective of PLS regression is to find a rank-$k$ decomposition of the data matrix $X$ into ``$X$-scores'' $T = X W \in \mathbb{R}^{n \times k}$ and ``$X$-loadings'' $P \in \mathbb{R}^{d \times k}$ so that $X \approx T P\tran = X W P\tran$, with weights $W \in \mathbb{R}^{d \times k}$.
The decomposition is subject to the constraint that the $X$-scores (or conversely: the corresponding $k$ components) can be used to predict $Y$, that is $Y \approx X W C\tran$, with ``$Y$-loadings'' $C \in \mathbb{R}^k$.
The $Y$-loadings can be seen as coefficients that quantify how much each of the $k$ components contributes to the prediction of the regression model.
There exist efficient methods for solving this optimization problem.\footnote{%
	We use the Scikit-learn \citep{pedregosa2011scikit} implementation of NIPALS \citep{wold1966estimation}.%
}

\begin{table}
	\centering
	\adjustbox{max width=\linewidth}{
		\begin{tabular}{llp{11em}lp{20em}lll}
\toprule
Property & Prop. ID & Entity & Entity ID & Prompt & Value & Unit\\
\midrule
birthyear & P569 & Nina Foch & Q235632 & In what year was Nina Foch born? & 1924 & annum\\
death year & P570 & Johannes R. Becher & Q58057 & In what year did Johannes R. Becher die? & 1958 & annum\\
population & P1082 & Akhisar & Q209905 & What is the population of Akhisar? & 173026 & 1\\
evelation & P2044 & Sondrio & Q6274 & How high is Sondrio? & 360 & metre\\
longitude & P625.long & Korean Empire & Q28233 & What is the longitude of Korean Empire? & 126.98 & degree\\
latitude & P625.lat & Küsnacht & Q69216 & What is the latitude of Küsnacht? & 47.32 & degree\\
\bottomrule
\end{tabular}

}
	\caption{Random sample of the entities used in our experiments, along with corresponding numeric attributes and prompts. See details and more samples in Appendix~\ref{sec:data_sample}.}
	\label{tbl:data_sample_small}
\end{table}

In our case $X$ corresponds to (a sample of) the activation space of a LM and $Y$ to corresponding LM outputs.
Concretely, for a given numeric property, such as birthyear, we collect $n$ entities that have this property.
For each entity $e$ we prepare a suitable prompt and encode this prompt with a LM to obtain an entity representation $x_e$ of dimension $d$.
That is, $X = \left[x_1 \: \cdots \: e_n \right]\tran \in \mathbb{R}^{n \times d}$.
Put simply, we encode prompts such as \emph{When was Karl Popper born?} and take the hidden state of a particular token at a particular layer as the LM's entity representation.
We also collect the quantity $y_e$ expressed by the LM when queried for the numeric attribute of entity $e$, that is, $Y = \{y_e \mid e \in \left[1 \twodots n \right]\} \subseteq \mathbb{R}^n$.
In the case of our running example query for Karl Popper's birthyear, Llama 2 expresses the quantity $y_e = 1902$.
Having collected entity representations $X$ and associated LM outputs $Y$, we fit a $k$-component PLS regression model to predict $Y$ from a $k$-dimensional subspace of $X$.

To measure how well numeric attributes can be predicted from low-dimensional subspaces of activation space, we vary the number of PLS regression components $k$, i.e., the dimensionality of the subspace, and record goodness of fit using the coefficient of determination $R^2 = 1-{\sum_{e=1}^{n}({y_e}-\hat{y_e})^2}/{\sum_{e=1}^{n}(y_e-\bar{y})^2}$, where $\hat{y_e}$ is the prediction of the regression model and $\bar{y}$ the sample-mean value of the numeric property in question.\footnote{%
	$R^2 = 1$ for a perfect model, $R^2 = 0$ for the mean (constant) model, and $R^2 < 0$ for models that perform worse than the mean model.}


\begin{figure}[t!]
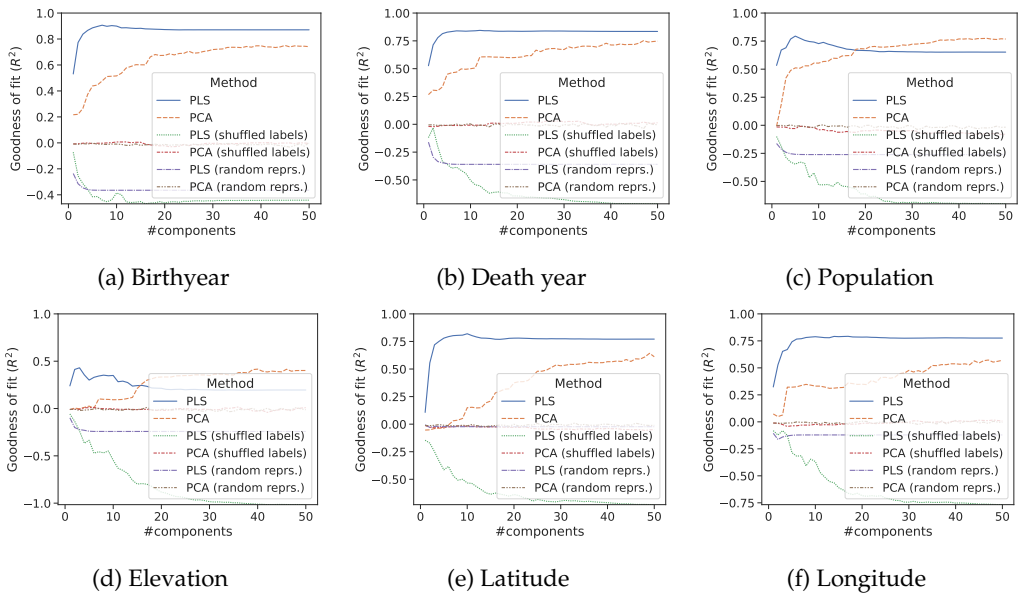

	\plssubfigs{Llama213bhf}
	\caption{Low-dimensional subspaces of \model{Llama-2-13B}'s 5120-dimensional activation space  are predictive of the quantity expressed by the LM when queried for a numeric attribute of an entity, across six different numeric properties.
	Each subfigure shows the performance of a regression model fitted to predict the expressed quantities from LM-internal entity representations (in layer $l=0.3$), as a function of the number of PCA/PLS components used for prediction.
	Unlike regression on PCA components (dashed orange), partial least squares regression (PLS, solid blue) identifies a small set of predictive components.
	Controls with shuffled labels (dotted green, dash-dotted red) and random entity representations (long-dash-dot purple, dash-dot-dot brown) fail to find predictive subspaces.}
	\label{fig:pls_results}
\end{figure}

\subsection{Results: Low-dimensional subspaces predictive of numeric attribute expression}
After selecting six of the most frequent numeric properties\footnote{%
These include longitude, which is non-monotonic due to the wrap-around at $\pm180^\circ$.
In defense of this choice we note that longitude is locally monotonic almost everywhere.}
in Wikidata \citep{vrandevcic2014wikidata}, for each property we randomly sample $n = 1000$ popular\footnote{We define popular entities as those in the top decile of the rank mean of Wikidata node degree and Wikipedia article length.} entities and prompt the LM (in English) for the corresponding attribute.
Samples of entities and prompts are shown in Table~\ref{tbl:data_sample_small} and Appendix~\ref{sec:data_sample}.
To obtain entity representations we take the $d$-dimensional hidden state of the entity mention's last subword token at layer $l$ to obtain a sample of the activation space $X$, choosing $l$ on a development set as described in Appendix~\ref{sec:edit_locus}.
We record the $n$ quantities parsed from the LM output to obtain $Y$ and fit PLS models with varying numbers of components to find property-encoding directions.

\begin{figure}[t!]
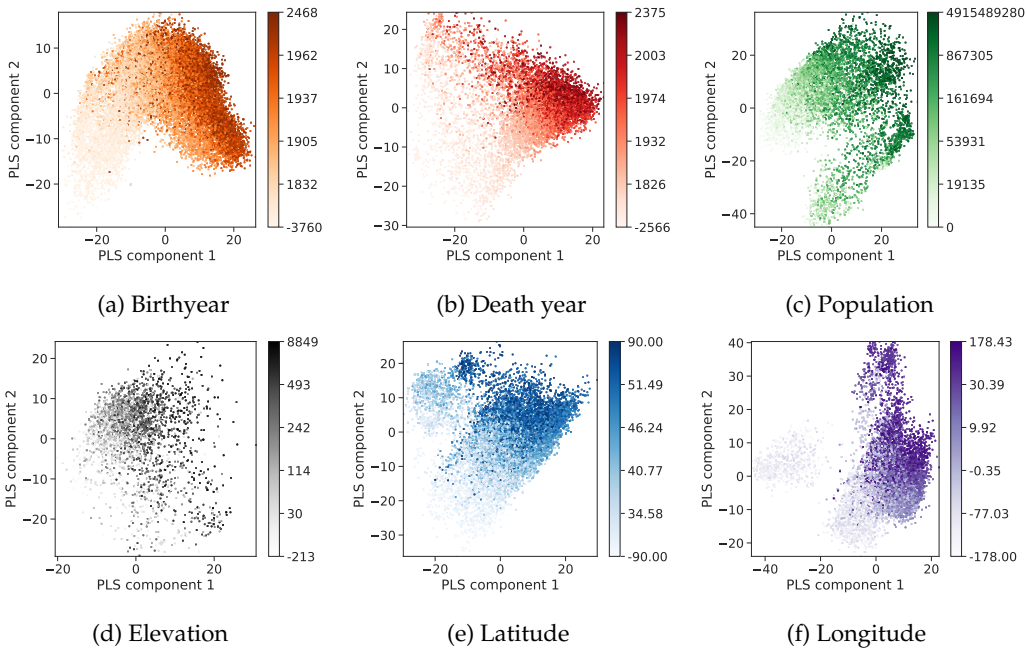

	\dimredsubfigs{Llama213bhf}
	\caption{Projection onto the top two components of per-property partial least squares regressions reveals monotonic structure in LM representations.
	We first fit a PLS model on \llamamed entity representations from our training split for each property, project entity representations from the test split, and then plot the resulting 2-d projections.
	Each dot represents one entity and color saturation represents the value of the corresponding entity attribute.
	See units for each property in Table~\ref{tbl:data_sample_small}.
	}
	\label{fig:dim_red}
\end{figure}

PLS regression results for \llamamed representations are shown in Figure~\ref{fig:pls_results} and results for additional models in Appendix~\ref{sec:more_regression_results}.
All numeric properties can be predicted well ($R^2 \ge 0.79$), with the exception of the elevation property ($R^2 = 0.43$).
Across all six properties, PLS identifies small sets of predictive components.
For example, a PLS model with $k=7$ components achieves a goodness of fit of $R^2 = 0.91$ when predicting birthyear attributes from entity representations.
Generally, all LMs appear to encode almost the entirety ($95\%$ of maximum $R^2$) of their stored numeric attribute information in two- to six-dimensional subspaces and a large fraction ($80\%$ of maximum $R^2$) appears to be encoded in two to three dimensions (see Appendix~\ref{sec:dim_reduction_data}).

To further illustrate the low dimensionality of numeric property representation, in Figure~\ref{fig:dim_red} we plot the projection of entity representations onto the top two components of PLS regressions.
The plots for each property exhibit monotonic structure.
Concomitant with poor regression fit, monotonic structure is least visible for the elevation property.
All other plots show clearly visible directions along which attribute values increase, reflecting the good fit of low-dimensional PLS regression models for these properties.

\section{Causal Effect of Property-Encoding Directions}
\label{sec:causation}

\subsection{Motivation: Do property-encoding directions affect model output?}
So far, we have found correlative evidence for the existence of directions in activation space that monotonically encode numeric properties.
Concretely, partial least squares regression on entity representations found directions that are predictive of what quantity the LM will express when queried for a particular numeric attribute of a particular entity.
However, representation is not a sufficient criterion for computation \citep{lasri2022probing}.
In our case this means that numeric properties might be encoded in representations without affecting model output.
In order to make the stronger claim that numeric properties are not only encoded monotonically, but that these representations have a monotonic effect on LM output, we now perform interventions to establish causality.

Intuitively, we want to find out if making ``small'' interventions leads to small changes in model output, if ``large'' interventions lead to large changes, and if the sign of the intervention matches the sign of the change.
We now formalize this intuition by adapting the definition of linear representation proposed by \citet{park2023linear} and \citet{jiang2024origins}.

\begin{definition}[Linear representation of numeric properties, adapted from \citet{jiang2024origins}]
	A numeric property is represented linearly if for all pairs of attribute instances $i, j$ with quantities $q_i \ne q_j$ and their representations $\vec{x_i}, \vec{x_j}$, there exists a \emph{steering vector} $\vec{u}$ so that $\vec{x_i} - \vec{x_j} \in \textrm{Cone}(\vec{u})$, where $\textrm{Cone}(\vec{v})=\left\{\alpha\vec{v}:\alpha>0\right\}$ is the cone of vector $\vec{v}$, i.e., its positive span.
\end{definition}

Linearity of representations only requires that representations lie in a cone, but says nothing about their ordering.
To model the natural structure of numeric properties, we introduce the constraint that the ordering of quantities is preserved in representation space.

\begin{definition}[Monotonic representation of numeric properties]
	A numeric property is represented monotonically if it is represented linearly in $\textrm{Cone}(\vec{u})$ and for all triples of attribute instances $h, i, j$ with quantities $q_h > q_i > q_j$ and representations $\vec{x_h}, \vec{x_i}, \vec{x_j}$ the following holds: $\vec{x_h}-\vec{x_j} = \alpha_{hj} \vec{u}$ and $\vec{x_i}-\vec{x_j} = \alpha_{ij} \vec{u}$ if and only if $\alpha_{hj} > \alpha_{ij}$.
\end{definition}
The coefficients $\alpha_{ij}$ and $\alpha_{hj}$ exist since the definition stipulates linear representation.
There are many ways to operationalize this definition.
One way is to prepare a ``synthetic'' series of monotonic representations in $\textrm{Cone}(\vec{u})$ by varying $\alpha$ and then testing if these representations result in monotonic output changes, which is what we will do now.

\subsection{Method: Activation patching}
Viewing the LM computation graph as causal graph \citep{meng2022locating,mcgrath2023hydra}, we intervene on model activations via activation patching \citep{vig2020investigating,wang2022interpretability,zhang2024best} and observe the effect on model output.
Unlike the common activation patching setup in which one replaces activations resulting from one input with activations from a different input, we create patches by editing activations along particular directions, similar to the activation manipulation method of \citet{matsumoto2022tracing}.

Specifically, for each of the top $K$ directions $\vec{u_k} \in R^d, k \in [1 \twodots K]$ found by PLS as described in \secref{sec:correlation}, we prepare patches $\vec{p}_{s,k} = \alpha_s \vec{u_k}$ with edit weights $\alpha_s$ and edit step index $s \in [1 \twodots S]$.\footnote{%
We choose $S = 80$ edits steps to balance step size and computational cost.}
In the view proposed by \citet{park2023linear} and \citet{jiang2024origins}, we can interpret PLS components as potential steering vectors and edit weights as their coefficients.
Lacking a principled method for choosing edit weights $\alpha_s$, we set their range to the minimum and maximum PLS loadings on each property's training split.
This choice yields patches covering the full empirical range of activation projections onto direction $\vec{u_k}$.
After sampling $n_{train} = 1000$ popular entities for each of the six numeric properties 
 we first fit PLS models for each property, then apply activation patches $\vec{p}_{s,k}$ to the representations of $n_{test} = 100$ held-out\footnote{%
 The held-out entities are entities which were not used to fit PLS models.}
 entities and for each entitiy record the LM's expressed quantity $y_{s,k}$.
To evaluate monotonicity, i.e., the notion that small (large) edit weights $\alpha_s,k$ should have a small (large) effects and that negative (positive) weights should decrease (increase) the expressed quantity $y_{s,k}$, we quantify the intervention effect via the ranked Spearman correlation $\rho(\alpha_{s,k};y_{s,k})$.

A question left open so far is where activation patching should be performed.
While automatic methods for localizing model components and subnetworks of interest have been proposed \citep{conmy2023automated,kramar2024atp}, for simplicity we perform a coarse, non-exhaustive search across layers and token positions based on one numeric property and use the found setting for all experiments (results shown in Appendix~\ref{sec:edit_locus}).
In addition to this edit locus, we also search for an edit window, whose purpose is to counteract self-repair \citep{mcgrath2023hydra} and iterative inference effects \citep{rushing2024explorations}.
Layer-wise we find that a window of $\pm2$ layers around the edit locus is most effective, which is smaller than the $\pm5$ layers used in prior work \citep{meng2022locating,hase2023does}.
We also implement a token-wise window \citep{monea2024glitch}, finding that in addition to the last entity mention token, patching up to two token representations to the left and one token representation to the right works best for the prompts in our experiments.
Typically, this token window size covers the entity mention and the main verb or last token of the prompt, depending on the numeric property (see prompts in Appendix~\ref{sec:data_sample}).
In summary, we patch activations in a 5-layer window centered on layer $l = 0.3$ and an up-to 4-token window surrounding the last entity mention subword token.
To improve output format adherence, we append the instruction \emph{One word answer only} to all prompts.

\subsection{Results: Property-encoding directions have effects and side-effects}

\begin{table}
	\centering
	\subcaptionbox{Birthyear of Karl Popper\label{tbl:edit_effect_birthyear}}{%
		\small
		\begin{tabular}{rrrrrrr}
\toprule
$\alpha_s$ & $y_{s,1}$ & $y_{s,2}$ & $y_{s,3}$ & $y_{s,4}$ & $y_{s,5}$ & $y_{s,6}$\\
\midrule
\textcolor[rgb]{0.71, 0.02, 0.15}{1.00} & \textcolor[rgb]{0.97, 0.74, 0.63}{1941} & \textcolor[rgb]{0.97, 0.66, 0.54}{1955} & \textcolor[rgb]{0.92, 0.50, 0.39}{1980} & \textcolor[rgb]{0.92, 0.50, 0.39}{1980} & \textcolor[rgb]{0.80, 0.24, 0.22}{2012} & \textcolor[rgb]{0.95, 0.79, 0.70}{1929}\\
\textcolor[rgb]{0.77, 0.20, 0.20}{0.90} & \textcolor[rgb]{0.97, 0.74, 0.63}{1941} & \textcolor[rgb]{0.97, 0.66, 0.54}{1955} & \textcolor[rgb]{0.97, 0.66, 0.54}{1955} & \textcolor[rgb]{0.91, 0.47, 0.37}{1984} & \textcolor[rgb]{0.80, 0.24, 0.22}{2012} & \textcolor[rgb]{0.95, 0.79, 0.70}{1929}\\
\textcolor[rgb]{0.84, 0.32, 0.26}{0.80} & \textcolor[rgb]{0.97, 0.74, 0.63}{1941} & \textcolor[rgb]{0.97, 0.66, 0.54}{1955} & \textcolor[rgb]{0.97, 0.66, 0.54}{1955} & \textcolor[rgb]{0.91, 0.47, 0.37}{1984} & \textcolor[rgb]{0.80, 0.24, 0.22}{2012} & \textcolor[rgb]{0.95, 0.79, 0.70}{1929}\\
\textcolor[rgb]{0.89, 0.43, 0.33}{0.70} & \textcolor[rgb]{0.97, 0.74, 0.63}{1941} & \textcolor[rgb]{0.97, 0.66, 0.54}{1955} & \textcolor[rgb]{0.97, 0.66, 0.54}{1955} & \textcolor[rgb]{0.92, 0.50, 0.39}{1980} & \textcolor[rgb]{0.95, 0.59, 0.47}{1968} & \textcolor[rgb]{0.95, 0.79, 0.70}{1929}\\
\textcolor[rgb]{0.93, 0.52, 0.41}{0.60} & \textcolor[rgb]{0.96, 0.78, 0.69}{1932} & \textcolor[rgb]{0.97, 0.66, 0.54}{1955} & \textcolor[rgb]{0.96, 0.76, 0.67}{1935} & \textcolor[rgb]{0.97, 0.65, 0.53}{1958} & \textcolor[rgb]{0.95, 0.59, 0.47}{1968} & \textcolor[rgb]{0.95, 0.79, 0.70}{1929}\\
\textcolor[rgb]{0.96, 0.60, 0.48}{0.50} & \textcolor[rgb]{0.96, 0.78, 0.69}{1932} & \textcolor[rgb]{0.97, 0.74, 0.64}{1940} & \textcolor[rgb]{0.96, 0.76, 0.67}{1935} & \textcolor[rgb]{0.97, 0.65, 0.53}{1958} & \textcolor[rgb]{0.96, 0.61, 0.49}{1964} & \textcolor[rgb]{0.95, 0.79, 0.70}{1929}\\
\textcolor[rgb]{0.97, 0.67, 0.56}{0.40} & \textcolor[rgb]{0.96, 0.78, 0.69}{1932} & \textcolor[rgb]{0.95, 0.78, 0.70}{1930} & \textcolor[rgb]{0.92, 0.83, 0.78}{1917} & \textcolor[rgb]{0.97, 0.65, 0.53}{1958} & \textcolor[rgb]{0.97, 0.65, 0.53}{1957} & \textcolor[rgb]{0.00, 0.00, 0.00}{1902}\\
\textcolor[rgb]{0.97, 0.74, 0.64}{0.30} & \textcolor[rgb]{0.95, 0.79, 0.70}{1929} & \textcolor[rgb]{0.95, 0.78, 0.70}{1930} & \textcolor[rgb]{0.88, 0.86, 0.84}{1906} & \textcolor[rgb]{0.97, 0.65, 0.53}{1958} & \textcolor[rgb]{0.95, 0.79, 0.70}{1929} & \textcolor[rgb]{0.00, 0.00, 0.00}{1902}\\
\textcolor[rgb]{0.95, 0.79, 0.72}{0.20} & \textcolor[rgb]{0.00, 0.00, 0.00}{1902} & \textcolor[rgb]{0.00, 0.00, 0.00}{1902} & \textcolor[rgb]{0.00, 0.00, 0.00}{1902} & \textcolor[rgb]{0.96, 0.77, 0.67}{1934} & \textcolor[rgb]{0.95, 0.79, 0.70}{1929} & \textcolor[rgb]{0.00, 0.00, 0.00}{1902}\\
\textcolor[rgb]{0.91, 0.84, 0.79}{0.10} & \textcolor[rgb]{0.00, 0.00, 0.00}{1902} & \textcolor[rgb]{0.00, 0.00, 0.00}{1902} & \textcolor[rgb]{0.00, 0.00, 0.00}{1902} & \textcolor[rgb]{0.00, 0.00, 0.00}{1902} & \textcolor[rgb]{0.00, 0.00, 0.00}{1902} & \textcolor[rgb]{0.00, 0.00, 0.00}{1902}\\
\textcolor[rgb]{0.00, 0.00, 0.00}{0.00} & \textcolor[rgb]{0.00, 0.00, 0.00}{1902} & \textcolor[rgb]{0.00, 0.00, 0.00}{1902} & \textcolor[rgb]{0.00, 0.00, 0.00}{1902} & \textcolor[rgb]{0.00, 0.00, 0.00}{1902} & \textcolor[rgb]{0.00, 0.00, 0.00}{1902} & \textcolor[rgb]{0.00, 0.00, 0.00}{1902}\\
\textcolor[rgb]{0.81, 0.85, 0.92}{-0.10} & \textcolor[rgb]{0.80, 0.85, 0.93}{1887} & \textcolor[rgb]{0.00, 0.00, 0.00}{1902} & \textcolor[rgb]{0.00, 0.00, 0.00}{1902} & \textcolor[rgb]{0.00, 0.00, 0.00}{1902} & \textcolor[rgb]{0.78, 0.84, 0.95}{1882} & \textcolor[rgb]{0.00, 0.00, 0.00}{1902}\\
\textcolor[rgb]{0.75, 0.83, 0.96}{-0.20} & \textcolor[rgb]{0.78, 0.84, 0.95}{1882} & \textcolor[rgb]{0.00, 0.00, 0.00}{1902} & \textcolor[rgb]{0.00, 0.00, 0.00}{1902} & \textcolor[rgb]{0.80, 0.85, 0.93}{1887} & \textcolor[rgb]{0.78, 0.84, 0.95}{1882} & \textcolor[rgb]{0.00, 0.00, 0.00}{1902}\\
\textcolor[rgb]{0.69, 0.79, 0.99}{-0.30} & \textcolor[rgb]{0.78, 0.84, 0.94}{1883} & \textcolor[rgb]{0.00, 0.00, 0.00}{1902} & \textcolor[rgb]{0.00, 0.00, 0.00}{1902} & \textcolor[rgb]{0.80, 0.85, 0.93}{1887} & \textcolor[rgb]{0.78, 0.84, 0.95}{1882} & \textcolor[rgb]{0.00, 0.00, 0.00}{1902}\\
\textcolor[rgb]{0.62, 0.74, 1.00}{-0.40} & \textcolor[rgb]{0.23, 0.30, 0.75}{1619} & \textcolor[rgb]{0.00, 0.00, 0.00}{1902} & \textcolor[rgb]{0.88, 0.86, 0.84}{1906} & \textcolor[rgb]{0.80, 0.85, 0.93}{1887} & \textcolor[rgb]{0.78, 0.84, 0.95}{1882} & \textcolor[rgb]{0.86, 0.86, 0.87}{1901}\\
\textcolor[rgb]{0.55, 0.69, 1.00}{-0.50} & \textcolor[rgb]{0.23, 0.30, 0.75}{1619} & \textcolor[rgb]{0.00, 0.00, 0.00}{1902} & \textcolor[rgb]{0.88, 0.86, 0.84}{1906} & \textcolor[rgb]{0.80, 0.85, 0.93}{1887} & \textcolor[rgb]{0.78, 0.84, 0.95}{1882} & \textcolor[rgb]{0.88, 0.86, 0.84}{1906}\\
\textcolor[rgb]{0.48, 0.62, 0.97}{-0.60} & \textcolor[rgb]{0.23, 0.30, 0.75}{1619} & \textcolor[rgb]{0.00, 0.00, 0.00}{1902} & \textcolor[rgb]{0.88, 0.86, 0.84}{1906} & \textcolor[rgb]{0.80, 0.85, 0.93}{1887} & \textcolor[rgb]{0.78, 0.84, 0.95}{1882} & \textcolor[rgb]{0.88, 0.86, 0.84}{1906}\\
\textcolor[rgb]{0.41, 0.55, 0.94}{-0.70} & \textcolor[rgb]{0.23, 0.30, 0.75}{1619} & \textcolor[rgb]{0.00, 0.00, 0.00}{1902} & \textcolor[rgb]{0.88, 0.86, 0.84}{1906} & \textcolor[rgb]{0.80, 0.85, 0.93}{1887} & \textcolor[rgb]{0.77, 0.84, 0.95}{1880} & \textcolor[rgb]{0.88, 0.86, 0.84}{1906}\\
\textcolor[rgb]{0.35, 0.47, 0.89}{-0.80} & \textcolor[rgb]{0.80, 0.85, 0.93}{1888} & \textcolor[rgb]{0.00, 0.00, 0.00}{1902} & \textcolor[rgb]{0.00, 0.00, 0.00}{1902} & \textcolor[rgb]{0.80, 0.85, 0.93}{1887} & \textcolor[rgb]{0.77, 0.84, 0.95}{1880} & \textcolor[rgb]{0.88, 0.86, 0.84}{1906}\\
\textcolor[rgb]{0.29, 0.38, 0.82}{-0.90} & \textcolor[rgb]{0.42, 0.55, 0.94}{1815} & \textcolor[rgb]{0.00, 0.00, 0.00}{1902} & \textcolor[rgb]{0.00, 0.00, 0.00}{1902} & \textcolor[rgb]{0.66, 0.77, 0.99}{1858} & \textcolor[rgb]{0.77, 0.84, 0.95}{1880} & \textcolor[rgb]{0.88, 0.86, 0.84}{1906}\\
\textcolor[rgb]{0.23, 0.30, 0.75}{-1.00} & \textcolor[rgb]{0.42, 0.55, 0.94}{1815} & \textcolor[rgb]{0.00, 0.00, 0.00}{1902} & \textcolor[rgb]{0.00, 0.00, 0.00}{1902} & \textcolor[rgb]{0.66, 0.77, 0.99}{1858} & \textcolor[rgb]{0.77, 0.84, 0.95}{1880} & \textcolor[rgb]{0.88, 0.86, 0.84}{1906}\\
\midrule
$\rho(\alpha_s, y_{s,k})$ & 0.91 & 0.87 & 0.72 & 0.97 & \textbf{0.98} & 0.39\\
\bottomrule
\end{tabular}

	}
	\qquad\qquad
	\subcaptionbox{Population of Zittau\label{tbl:edit_effect_population}}{%
		\small
		\begin{tabular}{rr}
\toprule
$\alpha_s$ & $y_{s,1}$\\
\midrule
\textcolor[rgb]{0.71, 0.02, 0.15}{1.00} & \textcolor[rgb]{0.71, 0.02, 0.15}{7.5 billion}\\
\textcolor[rgb]{0.77, 0.20, 0.20}{0.90} & \textcolor[rgb]{0.71, 0.02, 0.15}{7.5 billion}\\
\textcolor[rgb]{0.84, 0.32, 0.26}{0.80} & \textcolor[rgb]{0.71, 0.02, 0.15}{7.5 billion}\\
\textcolor[rgb]{0.89, 0.43, 0.33}{0.70} & \textcolor[rgb]{0.71, 0.02, 0.15}{7.5 billion}\\
\textcolor[rgb]{0.93, 0.52, 0.41}{0.60} & \textcolor[rgb]{0.71, 0.02, 0.15}{7.5 billion}\\
\textcolor[rgb]{0.96, 0.60, 0.48}{0.50} & \textcolor[rgb]{0.71, 0.02, 0.15}{7.5 billion}\\
\textcolor[rgb]{0.97, 0.67, 0.56}{0.40} & \textcolor[rgb]{0.71, 0.02, 0.15}{1.3 billion}\\
\textcolor[rgb]{0.97, 0.74, 0.64}{0.30} & \textcolor[rgb]{0.71, 0.02, 0.15}{1.3 billion}\\
\textcolor[rgb]{0.95, 0.79, 0.72}{0.20} & \textcolor[rgb]{0.71, 0.02, 0.15}{1.3 billion}\\
\textcolor[rgb]{0.91, 0.84, 0.79}{0.10} & \textcolor[rgb]{0.90, 0.85, 0.82}{10 million}\\
\textcolor[rgb]{0.00, 0.00, 0.00}{0.00} & \textcolor[rgb]{0.87, 0.86, 0.86}{40,000}\\
\textcolor[rgb]{0.81, 0.85, 0.92}{-0.10} & \textcolor[rgb]{0.87, 0.86, 0.86}{40,000}\\
\textcolor[rgb]{0.75, 0.83, 0.96}{-0.20} & \textcolor[rgb]{0.86, 0.87, 0.87}{25,000}\\
\textcolor[rgb]{0.69, 0.79, 0.99}{-0.30} & \textcolor[rgb]{0.86, 0.87, 0.87}{25,000}\\
\textcolor[rgb]{0.62, 0.74, 1.00}{-0.40} & \textcolor[rgb]{0.86, 0.87, 0.87}{20,000}\\
\textcolor[rgb]{0.55, 0.69, 1.00}{-0.50} & \textcolor[rgb]{0.86, 0.87, 0.87}{20,000}\\
\textcolor[rgb]{0.48, 0.62, 0.97}{-0.60} & \textcolor[rgb]{0.86, 0.87, 0.87}{20,000}\\
\textcolor[rgb]{0.41, 0.55, 0.94}{-0.70} & \textcolor[rgb]{0.86, 0.87, 0.87}{12,000}\\
\textcolor[rgb]{0.35, 0.47, 0.89}{-0.80} & \textcolor[rgb]{0.86, 0.87, 0.87}{12,000}\\
\textcolor[rgb]{0.29, 0.38, 0.82}{-0.90} & \textcolor[rgb]{0.86, 0.87, 0.87}{12,000}\\
\textcolor[rgb]{0.23, 0.30, 0.75}{-1.00} & \textcolor[rgb]{0.86, 0.87, 0.87}{12,000}\\
\midrule
$\rho(\alpha_s, y_{s,k})$ & \textbf{0.98}\\
\bottomrule
\end{tabular}

	}
	\caption{The quantity $y_{s,k}$ expressed by a LM changes as a result of directed activation patching along direction $k$ with (normalized) edit weight $\alpha_s$, with $\alpha_s = 0.00$ corresponding to unedited model activations.
	Warm colors indicate values larger than and cold colors values smaller than the true value, which, if output by the LM, is printed black.
	Table (\subref{tbl:edit_effect_birthyear}) shows how one-dimensional directed patches along each of the top six ``birthyear'' PLS components change the answer given by \llamamed to the prompt: \emph{In what year was Karl Popper born? One word answer only}.
	It is apparent that the most-correlated component ($k=1$) does not necessarily correspond to the direction in which model behavior exhibits highest monotonicity, which in this case is component $k=5$ with a Spearman correlation of $0.98$.
	Table (\subref{tbl:edit_effect_population}) shows the effect of patching along the top ``population'' component on \llamamed when prompted: \emph{What is the population of Zittau? One word answer only}.
	}
	\label{tbl:edit_effect_examples}
\end{table}

\begin{figure}[t!]
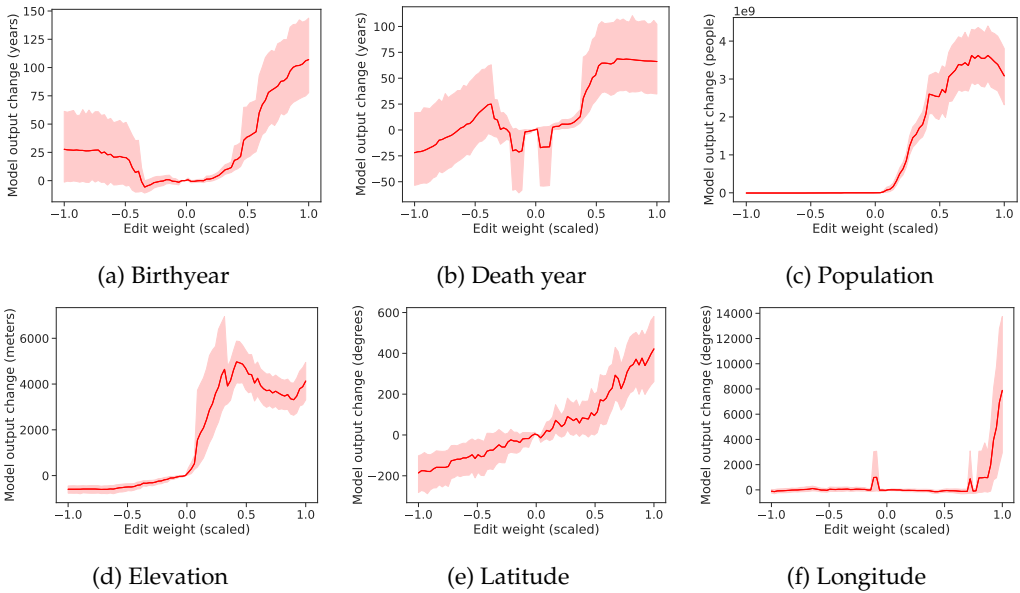

	\editsubfigs{llama213b}
	\caption{Effect of activation patching along property-specific directions across several numeric properties.
	Each subplot shows the change in the numeric attribute value expressed by \llamamed, as a function of the edit weight $\alpha_s$.
	Dark red lines indicate means across 100 entities sampled from held-out test sets and bands show standard deviations.
	}
	\label{fig:effects}
\end{figure}

We are interested in the effects and side effects on model output when patching activations along property-specific directions.
Looking at effects first, Table~\ref{tbl:edit_effect_examples} gives examples of how numeric attribute expression changes as a result of directed activation patching.
Patching along ``birthyear'' directions results in the expression of different years, although the degree of monotonicity, as quantified by Spearman correlation $\rho$, varies.
Patching along the top ``population'' direction causes the model to generate a range of outputs that can be interpreted as population sizes, although the largest values are more suited to a planetary than a municipal scale.
The sequence of outputs has rather sudden jumps, e.g., from \emph{40,000} (unedited model, $\alpha_s = 0.00$) to \emph{10 million} after taking the first step in the ``larger population'' direction ($\alpha_s = 0.10$).
The pattern of jumps and plateaus is plausibly connected to several factors such as tokenization effects and the likely high frequency of certain numerals (\emph{1.3 billion}: population of China at some point in time; \emph{7.5 billion}: population of Earth, etc.) in the training data, but we leave a detailed investigation to future work.
The pattern also indicates that activation space, while apparently monotonic, is not linear in this direction.
The intervention also induces a switch from positional notation (\emph{40,000}) to named numbers (\emph{million}, \emph{billion}), which showcases effects beyond single tokens.

Moving to a more systematic analysis, we plot mean-aggregated effects of directed activation patching across six numeric properties in Figure~\ref{fig:effects}.
We see that there are properties for which directed activation patching has highly monotonic effects, e.g., birthyear ($\rho=0.84$), elevation ($\rho=0.88$), or work period start ($\rho=0.90$), suggesting that these properties have highly monotonic representations.
Other properties exhibit a much smaller degree of monotonic editability, e.g., longitude ($\rho=0.55$) and population (0.65), suggesting that LM representations do not encode these properties as well.
Figures for additional models lead to similar to conclusions and are shown in Appendix~\ref{sec:more_edit_curves}.

\begin{figure}[t!]
	\centering
	\medsubfig{\llamasmall}{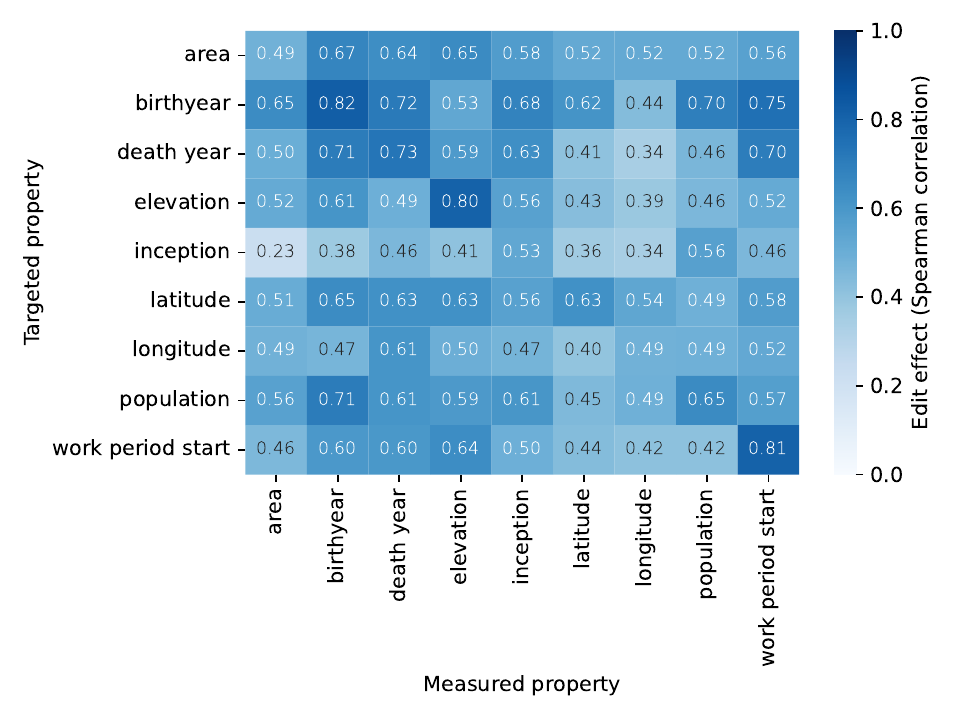}
	\medsubfig{\llamamed}{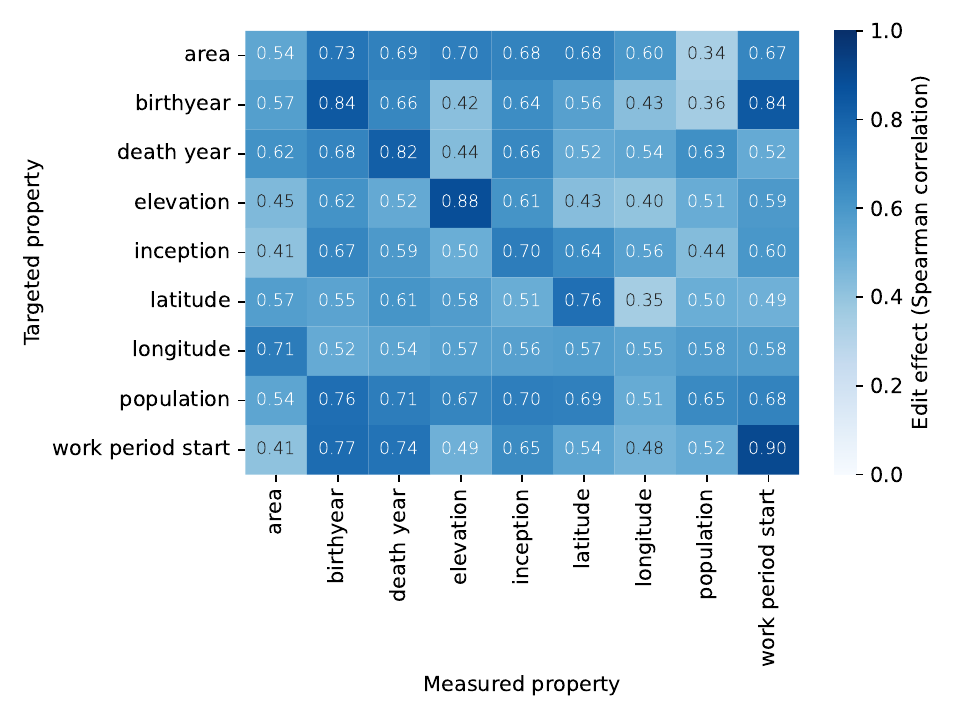}
	\caption{Mean-aggregated effects and side effects when performing activation patching along property-specific directions in activation space.
	Diagonal entries (top-left to bottom right) show the effect on the targeted property in terms of mean Spearman correlation between edit weight $alpha_s,k$ and expressed quantity $y_s,k$.
	For example, patching an entity representation along a ``birthyear'' direction results in a corresponding change in the quantity expressed by \llamamed with a correlation strength of $0.84$.
	Off-diagonal entries show the side-effects of activation patching, e.g., ``birthyear'' patches affect LM output when queried for an entity's death year with a correlation strength of $0.68$.}
	\label{fig:side_effects}
\end{figure}

Having observed the effects of our interventions we now turn to analyzing their side effects on the expression of unrelated, non-targeted numeric properties, i.e., properties that were not the target of the intervention.
For example, if we fitted a PLS regression to find and patch along ``birthyear'' directions, birthyear is our targeted property and all other properties, such as death year, population, or longitude are non-targeted properties.
Using the same directions found in \secref{sec:correlation}, we prompt LMs for non-targeted attributes, perform directed activation patching with weight $\alpha_s$ in a direction found for the targeted property and record expressed quantities $y'_{s,k}$.
To see if non-targeted properties are affected in a similar monotonic fashion as targeted ones, we quantify the side-effect of directed activation patching as the mean Spearman correlation $\rho(\alpha_s, y'_{s,k}$), aggregated over 100 entities per property.
We perform this procedure for all combinations of targeted and non-targeted properties, including three additional properties, and show results in Figure~\ref{fig:side_effects}.
In this figure, entries on the diagonal show the mean effect size for targeted properties and off-diagonal entries the size of side-effects.
For \llamasmall, the mean effect size $\bar{\rho}=0.65\pm0.12$ (i.e., the mean of diagonal entries), is not much larger than the mean side-effect size $\bar{\rho}=0.53\pm0.11$ (mean of off-diagonal entries).
In contrast, for \llamamed the effect size of $\bar{\rho}=0.85\pm0.07$ is considerably larger than the size of side effects ($\bar{\rho}=0.58\pm0.18$).
A plausible explanation for these results is that in case of \llamasmall different properties share a subspace which encodes generic numeric ranges or generic small-large ranges that are translated into quantities depending on context, while the representational space of \llamamed is more akin to a mixture of generic numeric and property-specific, ``more orthogonal'' subspaces.
Clearly, more work is needed to verify this hypothesis.

The analysis of side-effects is complicated by actual correlations between properties: Birthyear and death year distances are bounded by the human life span, latitude and population are correlated since the Earth's northern hemisphere is more populous, locations with higher elevation tend to have smaller populations, etc.
Consequently, one might argue that, say, editing an entity's birthyear should also affect LM output when querying the entity's death year.

\section{Related Work}
\label{sec:relwork}

Shaped by the locality of physical reality, the locality of human experience \citep{prystawski2023think} gives rise to distributional patterns of language use.
Such patterns include patterns of geographic and temporal coherence \citep{heinzerling2017trust}, which reflect spatiotemporal proximity of real-world entities.
These patterns can be picked up by statistical models and allow, e.g., to predict geographic information from co-occurrence statistics of cities mentioned in news articles \citep{louwerse2009language}.
Probing static word vector representations for numeric attributes of geopolitical entities, \citet{gupta2015distributional} obtain good relative rankings, but do not evaluate absolute values nor analyze the geometry of representations.
Continuing this line of research, \citet{lietard2021language} probe LM representations for GPS coordinates.
Perhaps due to the---by current standards---small scale of the studied LMs, they find only limited success but report that larger models appeared to encode more geographic information.
\citet{faisal2023geographic} measure how well the geographic proximity of countries can be recovered from LM representations but differ from our work in their focus on the impact of politico-cultural factors.

Closest to our work is the analysis of geo-temporal information encoded in Llama 2 representations by \citet{gurnee2023language}.
Our work corroborates their finding of linear subspaces of activation space which are predictive of numeric attributes, but is distinct in three important aspects.
First, as we show in \secref{sec:correlation}, the subspaces found PCA, as used by \citeauthor{gurnee2023language}, are of considerably higher dimensionality ($50-100$) than the subspaces found by partial least-square regression ($2-17$).
Our finding thus tightens the upper bound on the complexity of numeric property representation in recent LMs.
Second, we make explicit and formalize the notion of monotonic representation.
Third, our interventions via directed activation patching (\secref{sec:causation}) found one-dimensional directions with fine-grained effects on the expression of numeric attributes, across all numeric properties and models we analyzed, thereby establishing a causal relationship between monotonic representations and LM behavior.

\section{Limitations}
\label{sec:limitations}

\subsection{General limitations of representational analysis}

None of the language models studied in this work are embodied agents or otherwise capable of embodied cognition.
Lacking direct sensorimotor grounding \citep{harnad1990symbol,mollo2023vector,harnad2024language}, LMs cannot directly perceive, let alone precisely measure, the numerical attributes of which we claim to have found monotonic representations.
It follows that any such representations are an artifact of distributional patterns in their training data, and that the best one can hope for is isomorphy between model representations and the properties of the real-world entities to which we tie those representations.

Leaving the groundedness of representations aside, the idea that concepts, knowledge, or behavior are ``encoded'' in neural representations might seem intuitively appealing, but has been strongly critized, on theoretical grounds in the context of biological and artificial neural networks in general \citep{brette2019coding}, and on empirical grounds in the context of pretrained language models in particular \citep{hase2023does,niu2024knowledge}.

Analysis of LM representations also has well-known limitations.
Under the mild assumption that there exists a bijection between inputs and their representations, all information extractable from the input, i.e., the natural language prompt, can also be extracted from the LM's representation of that sequence \citep{pimentel2020information}.
Hence the question to be answered by representational analysis is not whether a feature of interest can be extracted or not, but how easy it is to extract.
How to best quantify ``ease of extraction'' \citep{pimentel2020information} is an open question, although methods have been proposed \citep{pimentel2020pareto,voita2020information}.

\subsection{Specific limitations of the representational analysis conducted in this work}

The low-dimensional linear subspaces found in this work allow relatively ``easy'' extraction when compared to the nominally high dimensionalities of activation space, but are still higher-dimensional than necessary, since the represented structures (e.g., years, geographic coordinates) are canonically one- to two-dimensional.
Furthermore, activation space is nominally high-dimensional but its intrinsic dimension is believed to be much lower \citep{li2018measuring,aghajanyan2021intrinsic,razzhigaev2024shape}.
For example \citet{razzhigaev2024shape} provide estimates for the intrinsic dimension of various LMs, ranging from about 10 to 70 dimensions (the models used in our experiments are not covered).
If we view a non-linear, non-monotonic representation of full intrinsic dimensionality as the most complex encoding with worst-case ease of extraction, and one- to two-dimensional linear monotonic encodings as the simplest representation with optimal ease of extraction, then the low-dimensional subspaces we found fall somewhere between these bounds.
Whether they are low-dimensional relative to the models' intrinsic dimension is currently unknown.
Put differently, if the intrinsic dimension of \llamasmall turns out to be, say, 10, then finding, a 10-dimensional subspace that encodes all latitude information (see Appendix~\ref{sec:dim_reduction_data}) is not surprising, but necessary.

While we found evidence for monotonic representation of numeric properties, it is likely that our causal interventions via activation patching along one-dimensional directions are too simplistic, considering the fact that according to our PLS regression results, numeric properties are encoded in low- but not one-dimensional subspaces.
Hence it is possible that a more refined editing method operating on higher-dimensional directions will allow more precise control over LM output.
Furthermore, our analysis is limited to popular entities, frequent numeric properties, and English queries, i.e., the combination most likely to be well-represented in the LM training data.

\section{Conclusions and Open Questions}
We used partial least-squares regression to identify low-dimensional subspaces of activation space that are predictive of the quantity an LM expresses when queried for numeric attributes such as an entity's birthyear.
We then performed activation patching along one-dimensional directions in these subspaces and observed corresponding changes in model output.
Our results suggest that LMs learn monotonic representations of numeric properties and that these monotonic representations exist in all of the language models we examined.

While this work takes a step towards a better understanding of numeric property representation in LMs, it leaves many questions unanswered.  Some of these are:
\begin{itemize}
	\item What exactly do the subspaces we found encode? Large side effects of activation patching suggest that there exist directions that encode generic numeric ranges. At the same time, the pronounced difference between large effect size and smaller side-effects we observed for the largest model in our analysis constitutes evidence for the existence of property-specific subspaces. Can we find more specific directions, possibly by adding constraints to minimize side effects into the regression objective, similar to \citep{meng2022locating}?
	\item If there are subspaces encoding generic numeric ranges, are changes along directions in these subspaces absolute or relative to typical value ranges? E.g., does a relative large change in birthyears, say 100 years, translate into a relatively large change in elevation, say 1000 meters, or into a change with the same value, i.e., 100 meters?
	\item Is the quality of numeric attribute representations in any way connected to LM performance on tasks involving the attributes in question? For example, does the model answer queries like \emph{Who was born earlier, Albert Einstein or Karl Popper?} correctly if the corresponding entity representations are appropriately positioned along a ``birthyear'' direction, and incorrectly if they are not?
\end{itemize}

\paragraph{Acknowledgements.}

This work was supported by JST CREST Grant Number JPMJCR20D2 and JSPS KAKENHI Grant Number 21K17814.

\bibliography{lit}

\begin{thebibliography}{71}
\providecommand{\natexlab}[1]{#1}
\providecommand{\url}[1]{\texttt{#1}}
\expandafter\ifx\csname urlstyle\endcsname\relax
  \providecommand{\doi}[1]{doi: #1}\else
  \providecommand{\doi}{doi: \begingroup \urlstyle{rm}\Url}\fi

\bibitem[Aghajanyan et~al.(2021)Aghajanyan, Gupta, and
  Zettlemoyer]{aghajanyan2021intrinsic}
Armen Aghajanyan, Sonal Gupta, and Luke Zettlemoyer.
\newblock Intrinsic dimensionality explains the effectiveness of language model
  fine-tuning.
\newblock In Chengqing Zong, Fei Xia, Wenjie Li, and Roberto Navigli (eds.),
  \emph{Proceedings of the 59th Annual Meeting of the Association for
  Computational Linguistics and the 11th International Joint Conference on
  Natural Language Processing (Volume 1: Long Papers)}, pp.\  7319--7328,
  Online, August 2021. Association for Computational Linguistics.
\newblock \doi{10.18653/v1/2021.acl-long.568}.
\newblock URL \url{https://aclanthology.org/2021.acl-long.568}.

\bibitem[Almazrouei et~al.(2023)Almazrouei, Alobeidli, Alshamsi, Cappelli,
  Cojocaru, Debbah, Étienne Goffinet, Hesslow, Launay, Malartic, Mazzotta,
  Noune, Pannier, and Penedo]{almazrouei2023falcon}
Ebtesam Almazrouei, Hamza Alobeidli, Abdulaziz Alshamsi, Alessandro Cappelli,
  Ruxandra Cojocaru, Mérouane Debbah, Étienne Goffinet, Daniel Hesslow,
  Julien Launay, Quentin Malartic, Daniele Mazzotta, Badreddine Noune, Baptiste
  Pannier, and Guilherme Penedo.
\newblock The falcon series of open language models, 2023.

\bibitem[Bender \& Koller(2020)Bender and Koller]{bender2020climbing}
Emily~M. Bender and Alexander Koller.
\newblock Climbing towards {NLU}: {On} meaning, form, and understanding in the
  age of data.
\newblock In Dan Jurafsky, Joyce Chai, Natalie Schluter, and Joel Tetreault
  (eds.), \emph{Proceedings of the 58th Annual Meeting of the Association for
  Computational Linguistics}, pp.\  5185--5198, Online, July 2020. Association
  for Computational Linguistics.
\newblock \doi{10.18653/v1/2020.acl-main.463}.
\newblock URL \url{https://aclanthology.org/2020.acl-main.463}.

\bibitem[Brette(2019)]{brette2019coding}
Romain Brette.
\newblock Is coding a relevant metaphor for the brain?
\newblock \emph{Behavioral and Brain Sciences}, 42:\penalty0 e215, 2019.

\bibitem[Conant \& Ashby(1970)Conant and Ashby]{conant1970every}
Roger~C. Conant and W.~Ross Ashby.
\newblock Every good regulator of a system must be a model of that system.
\newblock \emph{International journal of systems science}, 1\penalty0
  (2):\penalty0 89--97, 1970.

\bibitem[Conmy et~al.(2023)Conmy, Mavor-Parker, Lynch, Heimersheim, and
  Garriga-Alonso]{conmy2023automated}
Arthur Conmy, Augustine~N. Mavor-Parker, Aengus Lynch, Stefan Heimersheim, and
  Adrià Garriga-Alonso.
\newblock Towards automated circuit discovery for mechanistic interpretability,
  2023.

\bibitem[De~Cao et~al.(2021)De~Cao, Aziz, and Titov]{decao2021editing}
Nicola De~Cao, Wilker Aziz, and Ivan Titov.
\newblock Editing factual knowledge in language models.
\newblock In Marie-Francine Moens, Xuanjing Huang, Lucia Specia, and Scott
  Wen-tau Yih (eds.), \emph{Proceedings of the 2021 Conference on Empirical
  Methods in Natural Language Processing}, pp.\  6491--6506, Online and Punta
  Cana, Dominican Republic, November 2021. Association for Computational
  Linguistics.
\newblock \doi{10.18653/v1/2021.emnlp-main.522}.
\newblock URL \url{https://aclanthology.org/2021.emnlp-main.522}.

\bibitem[Elhage et~al.(2022)Elhage, Hume, Olsson, Schiefer, Henighan, Kravec,
  Hatfield-Dodds, Lasenby, Drain, Chen, Grosse, McCandlish, Kaplan, Amodei,
  Wattenberg, and Olah]{elhage2022toy}
Nelson Elhage, Tristan Hume, Catherine Olsson, Nicholas Schiefer, Tom Henighan,
  Shauna Kravec, Zac Hatfield-Dodds, Robert Lasenby, Dawn Drain, Carol Chen,
  Roger Grosse, Sam McCandlish, Jared Kaplan, Dario Amodei, Martin Wattenberg,
  and Christopher Olah.
\newblock Toy models of superposition, 2022.

\bibitem[Faisal \& Anastasopoulos(2023)Faisal and
  Anastasopoulos]{faisal2023geographic}
Fahim Faisal and Antonios Anastasopoulos.
\newblock Geographic and geopolitical biases of language models.
\newblock In Duygu Ataman (ed.), \emph{Proceedings of the 3rd Workshop on
  Multi-lingual Representation Learning (MRL)}, pp.\  139--163, Singapore,
  December 2023. Association for Computational Linguistics.
\newblock \doi{10.18653/v1/2023.mrl-1.12}.
\newblock URL \url{https://aclanthology.org/2023.mrl-1.12}.

\bibitem[Geva et~al.(2023)Geva, Bastings, Filippova, and
  Globerson]{geva2023dissecting}
Mor Geva, Jasmijn Bastings, Katja Filippova, and Amir Globerson.
\newblock Dissecting recall of factual associations in auto-regressive language
  models.
\newblock In Houda Bouamor, Juan Pino, and Kalika Bali (eds.),
  \emph{Proceedings of the 2023 Conference on Empirical Methods in Natural
  Language Processing}, pp.\  12216--12235, Singapore, December 2023.
  Association for Computational Linguistics.
\newblock \doi{10.18653/v1/2023.emnlp-main.751}.
\newblock URL \url{https://aclanthology.org/2023.emnlp-main.751}.

\bibitem[Godey et~al.(2024)Godey, Éric de~la Clergerie, and
  Sagot]{godey2024scaling}
Nathan Godey, Éric de~la Clergerie, and Benoît Sagot.
\newblock On the scaling laws of geographical representation in language
  models, 2024.

\bibitem[Gupta et~al.(2015)Gupta, Boleda, Baroni, and
  Pad{\'o}]{gupta2015distributional}
Abhijeet Gupta, Gemma Boleda, Marco Baroni, and Sebastian Pad{\'o}.
\newblock Distributional vectors encode referential attributes.
\newblock In Llu{\'\i}s M{\`a}rquez, Chris Callison-Burch, and Jian Su (eds.),
  \emph{Proceedings of the 2015 Conference on Empirical Methods in Natural
  Language Processing}, pp.\  12--21, Lisbon, Portugal, September 2015.
  Association for Computational Linguistics.
\newblock \doi{10.18653/v1/D15-1002}.
\newblock URL \url{https://aclanthology.org/D15-1002}.

\bibitem[Gurnee \& Tegmark(2023)Gurnee and Tegmark]{gurnee2023language}
Wes Gurnee and Max Tegmark.
\newblock Language models represent space and time, 2023.

\bibitem[Harnad(1990)]{harnad1990symbol}
Stevan Harnad.
\newblock The symbol grounding problem.
\newblock \emph{Physica D: Nonlinear Phenomena}, 42\penalty0 (1-3):\penalty0
  335--346, 1990.

\bibitem[Harnad(2024)]{harnad2024language}
Stevan Harnad.
\newblock Language writ large: Llms, chatgpt, grounding, meaning and
  understanding, 2024.

\bibitem[Harris et~al.(2020)Harris, Millman, van~der Walt, Gommers, Virtanen,
  Cournapeau, Wieser, Taylor, Berg, Smith, Kern, Picus, Hoyer, van Kerkwijk,
  Brett, Haldane, del R{\'{i}}o, Wiebe, Peterson, G{\'{e}}rard-Marchant,
  Sheppard, Reddy, Weckesser, Abbasi, Gohlke, and Oliphant]{harris2020array}
Charles~R. Harris, K.~Jarrod Millman, St{\'{e}}fan~J. van~der Walt, Ralf
  Gommers, Pauli Virtanen, David Cournapeau, Eric Wieser, Julian Taylor,
  Sebastian Berg, Nathaniel~J. Smith, Robert Kern, Matti Picus, Stephan Hoyer,
  Marten~H. van Kerkwijk, Matthew Brett, Allan Haldane, Jaime~Fern{\'{a}}ndez
  del R{\'{i}}o, Mark Wiebe, Pearu Peterson, Pierre G{\'{e}}rard-Marchant,
  Kevin Sheppard, Tyler Reddy, Warren Weckesser, Hameer Abbasi, Christoph
  Gohlke, and Travis~E. Oliphant.
\newblock Array programming with {NumPy}.
\newblock \emph{Nature}, 585\penalty0 (7825):\penalty0 357--362, September
  2020.
\newblock \doi{10.1038/s41586-020-2649-2}.
\newblock URL \url{https://doi.org/10.1038/s41586-020-2649-2}.

\bibitem[Hase et~al.(2023{\natexlab{a}})Hase, Bansal, Kim, and
  Ghandeharioun]{hase2023does}
Peter Hase, Mohit Bansal, Been Kim, and Asma Ghandeharioun.
\newblock Does localization inform editing? surprising differences in
  causality-based localization vs. knowledge editing in language models,
  2023{\natexlab{a}}.

\bibitem[Hase et~al.(2023{\natexlab{b}})Hase, Diab, Celikyilmaz, Li, Kozareva,
  Stoyanov, Bansal, and Iyer]{hase2023beliefs}
Peter Hase, Mona Diab, Asli Celikyilmaz, Xian Li, Zornitsa Kozareva, Veselin
  Stoyanov, Mohit Bansal, and Srinivasan Iyer.
\newblock Methods for measuring, updating, and visualizing factual beliefs in
  language models.
\newblock In \emph{Proceedings of the 17th Conference of the European Chapter
  of the Association for Computational Linguistics}, pp.\  2714--2731,
  Dubrovnik, Croatia, May 2023{\natexlab{b}}. Association for Computational
  Linguistics.
\newblock URL \url{https://aclanthology.org/2023.eacl-main.199}.

\bibitem[Heinzerling \& Inui(2021)Heinzerling and
  Inui]{heinzerling2021language}
Benjamin Heinzerling and Kentaro Inui.
\newblock Language models as knowledge bases: On entity representations,
  storage capacity, and paraphrased queries.
\newblock In Paola Merlo, Jorg Tiedemann, and Reut Tsarfaty (eds.),
  \emph{Proceedings of the 16th Conference of the European Chapter of the
  Association for Computational Linguistics: Main Volume}, pp.\  1772--1791,
  Online, April 2021. Association for Computational Linguistics.
\newblock \doi{10.18653/v1/2021.eacl-main.153}.
\newblock URL \url{https://aclanthology.org/2021.eacl-main.153}.

\bibitem[Heinzerling et~al.(2017)Heinzerling, Strube, and
  Lin]{heinzerling2017trust}
Benjamin Heinzerling, Michael Strube, and Chin-Yew Lin.
\newblock Trust, but verify! better entity linking through automatic
  verification.
\newblock In Mirella Lapata, Phil Blunsom, and Alexander Koller (eds.),
  \emph{Proceedings of the 15th Conference of the {E}uropean Chapter of the
  Association for Computational Linguistics: Volume 1, Long Papers}, pp.\
  828--838, Valencia, Spain, April 2017. Association for Computational
  Linguistics.
\newblock URL \url{https://aclanthology.org/E17-1078}.

\bibitem[Hernandez et~al.(2023)Hernandez, Li, and Andreas]{hernandez2023remedi}
Evan Hernandez, Belinda~Z. Li, and Jacob Andreas.
\newblock Inspecting and editing knowledge representations in language models.
\newblock In \emph{Arxiv}, 2023.
\newblock URL \url{https://arxiv.org/abs/2304.00740}.

\bibitem[Hunter(2007)]{hunter2007matplotlib}
J.~D. Hunter.
\newblock Matplotlib: A 2d graphics environment.
\newblock \emph{Computing in Science \& Engineering}, 9\penalty0 (3):\penalty0
  90--95, 2007.
\newblock \doi{10.1109/MCSE.2007.55}.

\bibitem[Jiang et~al.(2023)Jiang, Sablayrolles, Mensch, Bamford, Chaplot,
  de~las Casas, Bressand, Lengyel, Lample, Saulnier, Lavaud, Lachaux, Stock,
  Scao, Lavril, Wang, Lacroix, and Sayed]{jiang2023mistral}
Albert~Q. Jiang, Alexandre Sablayrolles, Arthur Mensch, Chris Bamford,
  Devendra~Singh Chaplot, Diego de~las Casas, Florian Bressand, Gianna Lengyel,
  Guillaume Lample, Lucile Saulnier, Lélio~Renard Lavaud, Marie-Anne Lachaux,
  Pierre Stock, Teven~Le Scao, Thibaut Lavril, Thomas Wang, Timothée Lacroix,
  and William~El Sayed.
\newblock Mistral 7b, 2023.

\bibitem[Jiang et~al.(2024)Jiang, Rajendran, Ravikumar, Aragam, and
  Veitch]{jiang2024origins}
Yibo Jiang, Goutham Rajendran, Pradeep Ravikumar, Bryon Aragam, and Victor
  Veitch.
\newblock On the origins of linear representations in large language models,
  2024.

\bibitem[Jiang et~al.(2020)Jiang, Xu, Araki, and Neubig]{jiang2020know}
Zhengbao Jiang, Frank~F. Xu, Jun Araki, and Graham Neubig.
\newblock How can we know what language models know?
\newblock \emph{Transactions of the Association for Computational Linguistics},
  8:\penalty0 423--438, 2020.
\newblock \doi{10.1162/tacl_a_00324}.
\newblock URL \url{https://aclanthology.org/2020.tacl-1.28}.

\bibitem[Jiang et~al.(2021)Jiang, Araki, Ding, and Neubig]{jiang2021know}
Zhengbao Jiang, Jun Araki, Haibo Ding, and Graham Neubig.
\newblock How can we know when language models know? on the calibration of
  language models for question answering.
\newblock \emph{Transactions of the Association for Computational Linguistics},
  9:\penalty0 962--977, 2021.
\newblock \doi{10.1162/tacl_a_00407}.
\newblock URL \url{https://aclanthology.org/2021.tacl-1.57}.

\bibitem[Kassner et~al.(2021)Kassner, Dufter, and
  Sch{\"u}tze]{kassner2021multilingual}
Nora Kassner, Philipp Dufter, and Hinrich Sch{\"u}tze.
\newblock Multilingual {LAMA}: Investigating knowledge in multilingual
  pretrained language models.
\newblock In Paola Merlo, Jorg Tiedemann, and Reut Tsarfaty (eds.),
  \emph{Proceedings of the 16th Conference of the European Chapter of the
  Association for Computational Linguistics: Main Volume}, pp.\  3250--3258,
  Online, April 2021. Association for Computational Linguistics.
\newblock \doi{10.18653/v1/2021.eacl-main.284}.
\newblock URL \url{https://aclanthology.org/2021.eacl-main.284}.

\bibitem[Kramár et~al.(2024)Kramár, Lieberum, Shah, and Nanda]{kramar2024atp}
János Kramár, Tom Lieberum, Rohin Shah, and Neel Nanda.
\newblock Atp*: An efficient and scalable method for localizing llm behaviour
  to components, 2024.

\bibitem[Lasri et~al.(2022)Lasri, Pimentel, Lenci, Poibeau, and
  Cotterell]{lasri2022probing}
Karim Lasri, Tiago Pimentel, Alessandro Lenci, Thierry Poibeau, and Ryan
  Cotterell.
\newblock Probing for the usage of grammatical number.
\newblock In Smaranda Muresan, Preslav Nakov, and Aline Villavicencio (eds.),
  \emph{Proceedings of the 60th Annual Meeting of the Association for
  Computational Linguistics (Volume 1: Long Papers)}, pp.\  8818--8831, Dublin,
  Ireland, May 2022. Association for Computational Linguistics.
\newblock \doi{10.18653/v1/2022.acl-long.603}.
\newblock URL \url{https://aclanthology.org/2022.acl-long.603}.

\bibitem[Lederman \& Mahowald(2024)Lederman and Mahowald]{lederman2024language}
Harvey Lederman and Kyle Mahowald.
\newblock Are language models more like libraries or like librarians?
  bibliotechnism, the novel reference problem, and the attitudes of llms, 2024.

\bibitem[Li et~al.(2018)Li, Farkhoor, Liu, and Yosinski]{li2018measuring}
Chunyuan Li, Heerad Farkhoor, Rosanne Liu, and Jason Yosinski.
\newblock Measuring the intrinsic dimension of objective landscapes.
\newblock \emph{arXiv preprint arXiv:1804.08838}, 2018.

\bibitem[Li{\'e}tard et~al.(2021)Li{\'e}tard, Abdou, and
  S{\o}gaard]{lietard2021language}
Bastien Li{\'e}tard, Mostafa Abdou, and Anders S{\o}gaard.
\newblock Do language models know the way to {R}ome?
\newblock In Jasmijn Bastings, Yonatan Belinkov, Emmanuel Dupoux, Mario
  Giulianelli, Dieuwke Hupkes, Yuval Pinter, and Hassan Sajjad (eds.),
  \emph{Proceedings of the Fourth BlackboxNLP Workshop on Analyzing and
  Interpreting Neural Networks for NLP}, pp.\  510--517, Punta Cana, Dominican
  Republic, November 2021. Association for Computational Linguistics.
\newblock \doi{10.18653/v1/2021.blackboxnlp-1.40}.
\newblock URL \url{https://aclanthology.org/2021.blackboxnlp-1.40}.

\bibitem[Liu et~al.(2022)Liu, Kitouni, Nolte, Michaud, Tegmark, and
  Williams]{liu2022grokking}
Ziming Liu, Ouail Kitouni, Niklas~S Nolte, Eric Michaud, Max Tegmark, and Mike
  Williams.
\newblock Towards understanding grokking: An effective theory of representation
  learning.
\newblock In S.~Koyejo, S.~Mohamed, A.~Agarwal, D.~Belgrave, K.~Cho, and A.~Oh
  (eds.), \emph{Advances in Neural Information Processing Systems}, volume~35,
  pp.\  34651--34663. Curran Associates, Inc., 2022.
\newblock URL
  \url{https://proceedings.neurips.cc/paper_files/paper/2022/file/dfc310e81992d2e4cedc09ac47eff13e-Paper-Conference.pdf}.

\bibitem[Louwerse \& Zwaan(2009)Louwerse and Zwaan]{louwerse2009language}
Max~M. Louwerse and Rolf~A. Zwaan.
\newblock Language encodes geographical information.
\newblock \emph{Cognitive Science}, 33\penalty0 (1):\penalty0 51--73, 2009.
\newblock \doi{https://doi.org/10.1111/j.1551-6709.2008.01003.x}.

\bibitem[Marks \& Tegmark(2023)Marks and Tegmark]{marks2023geometry}
Samuel Marks and Max Tegmark.
\newblock The geometry of truth: Emergent linear structure in large language
  model representations of true/false datasets, 2023.

\bibitem[Matsumoto et~al.(2022)Matsumoto, Heinzerling, Yoshikawa, and
  Inui]{matsumoto2022tracing}
Yuta Matsumoto, Benjamin Heinzerling, Masashi Yoshikawa, and Kentaro Inui.
\newblock Tracing and manipulating intermediate values in neural math problem
  solvers.
\newblock In Deborah Ferreira, Marco Valentino, Andre Freitas, Sean Welleck,
  and Moritz Schubotz (eds.), \emph{Proceedings of the 1st Workshop on
  Mathematical Natural Language Processing (MathNLP)}, pp.\  1--6, Abu Dhabi,
  United Arab Emirates (Hybrid), December 2022. Association for Computational
  Linguistics.
\newblock \doi{10.18653/v1/2022.mathnlp-1.1}.
\newblock URL \url{https://aclanthology.org/2022.mathnlp-1.1}.

\bibitem[McGrath et~al.(2023)McGrath, Rahtz, Kramar, Mikulik, and
  Legg]{mcgrath2023hydra}
Thomas McGrath, Matthew Rahtz, Janos Kramar, Vladimir Mikulik, and Shane Legg.
\newblock The hydra effect: Emergent self-repair in language model
  computations, 2023.

\bibitem[Meng et~al.(2022)Meng, Bau, Andonian, and Belinkov]{meng2022locating}
Kevin Meng, David Bau, Alex Andonian, and Yonatan Belinkov.
\newblock Locating and editing factual associations in {GPT}.
\newblock \emph{Advances in Neural Information Processing Systems}, 36, 2022.

\bibitem[Merullo et~al.(2023)Merullo, Eickhoff, and
  Pavlick]{merullo2023mechanism}
Jack Merullo, Carsten Eickhoff, and Ellie Pavlick.
\newblock A mechanism for solving relational tasks in transformer language
  models, 2023.

\bibitem[Mitchell et~al.(2021)Mitchell, Lin, Bosselut, Finn, and
  Manning]{mitchell2021fast}
Eric Mitchell, Charles Lin, Antoine Bosselut, Chelsea Finn, and Christopher~D.
  Manning.
\newblock Fast model editing at scale.
\newblock \emph{CoRR}, 2021.
\newblock URL \url{https://arxiv.org/pdf/2110.11309.pdf}.

\bibitem[Mollo \& Millière(2023)Mollo and Millière]{mollo2023vector}
Dimitri~Coelho Mollo and Raphaël Millière.
\newblock The vector grounding problem, 2023.

\bibitem[Monea et~al.(2024)Monea, Peyrard, Josifoski, Chaudhary, Eisner,
  Kıcıman, Palangi, Patra, and West]{monea2024glitch}
Giovanni Monea, Maxime Peyrard, Martin Josifoski, Vishrav Chaudhary, Jason
  Eisner, Emre Kıcıman, Hamid Palangi, Barun Patra, and Robert West.
\newblock A glitch in the matrix? locating and detecting language model
  grounding with fakepedia, 2024.

\bibitem[Nanda et~al.(2023)Nanda, Lee, and Wattenberg]{nanda2023emergent}
Neel Nanda, Andrew Lee, and Martin Wattenberg.
\newblock Emergent linear representations in world models of self-supervised
  sequence models.
\newblock In Yonatan Belinkov, Sophie Hao, Jaap Jumelet, Najoung Kim, Arya
  McCarthy, and Hosein Mohebbi (eds.), \emph{Proceedings of the 6th BlackboxNLP
  Workshop: Analyzing and Interpreting Neural Networks for NLP}, pp.\  16--30,
  Singapore, December 2023. Association for Computational Linguistics.
\newblock \doi{10.18653/v1/2023.blackboxnlp-1.2}.
\newblock URL \url{https://aclanthology.org/2023.blackboxnlp-1.2}.

\bibitem[Niu et~al.(2024)Niu, Liu, Zhu, and Penn]{niu2024knowledge}
Jingcheng Niu, Andrew Liu, Zining Zhu, and Gerald Penn.
\newblock What does the knowledge neuron thesis have to do with knowledge?
\newblock In \emph{The Twelfth International Conference on Learning
  Representations}, 2024.
\newblock URL \url{https://openreview.net/forum?id=2HJRwwbV3G}.

\bibitem[{P}andas~development team(2020)]{reback2020pandas}
The {P}andas~development team.
\newblock pandas-dev/pandas: Pandas, February 2020.
\newblock URL \url{https://doi.org/10.5281/zenodo.3509134}.

\bibitem[Park et~al.(2023)Park, Choe, and Veitch]{park2023linear}
Kiho Park, Yo~Joong Choe, and Victor Veitch.
\newblock The linear representation hypothesis and the geometry of large
  language models, 2023.

\bibitem[Paszke et~al.(2019)Paszke, Gross, Massa, Lerer, Bradbury, Chanan,
  Killeen, Lin, Gimelshein, Antiga, et~al.]{paszke2019pytorch}
Adam Paszke, Sam Gross, Francisco Massa, Adam Lerer, James Bradbury, Gregory
  Chanan, Trevor Killeen, Zeming Lin, Natalia Gimelshein, Luca Antiga, et~al.
\newblock Pytorch: An imperative style, high-performance deep learning library.
\newblock \emph{Advances in neural information processing systems}, 32, 2019.

\bibitem[Pearson(1901)]{pearson1901lines}
Karl Pearson.
\newblock On lines and planes of closest fit to systems of points in space.
\newblock \emph{The London, Edinburgh, and Dublin Philosophical Magazine and
  Journal of Science}, 2\penalty0 (11):\penalty0 559--572, 1901.
\newblock \doi{10.1080/14786440109462720}.
\newblock URL \url{https://doi.org/10.1080/14786440109462720}.

\bibitem[Pedregosa et~al.(2011)Pedregosa, Varoquaux, Gramfort, Michel, Thirion,
  Grisel, Blondel, Prettenhofer, Weiss, Dubourg, Vanderplas, Passos,
  Cournapeau, Brucher, Perrot, and Duchesnay]{pedregosa2011scikit}
F.~Pedregosa, G.~Varoquaux, A.~Gramfort, V.~Michel, B.~Thirion, O.~Grisel,
  M.~Blondel, P.~Prettenhofer, R.~Weiss, V.~Dubourg, J.~Vanderplas, A.~Passos,
  D.~Cournapeau, M.~Brucher, M.~Perrot, and E.~Duchesnay.
\newblock Scikit-learn: Machine learning in {P}ython.
\newblock \emph{Journal of Machine Learning Research}, 12:\penalty0 2825--2830,
  2011.

\bibitem[Petroni et~al.(2019)Petroni, Rockt{\"a}schel, Riedel, Lewis, Bakhtin,
  Wu, and Miller]{petroni2019language}
Fabio Petroni, Tim Rockt{\"a}schel, Sebastian Riedel, Patrick Lewis, Anton
  Bakhtin, Yuxiang Wu, and Alexander Miller.
\newblock Language models as knowledge bases?
\newblock In Kentaro Inui, Jing Jiang, Vincent Ng, and Xiaojun Wan (eds.),
  \emph{Proceedings of the 2019 Conference on Empirical Methods in Natural
  Language Processing and the 9th International Joint Conference on Natural
  Language Processing (EMNLP-IJCNLP)}, pp.\  2463--2473, Hong Kong, China,
  November 2019. Association for Computational Linguistics.
\newblock \doi{10.18653/v1/D19-1250}.
\newblock URL \url{https://aclanthology.org/D19-1250}.

\bibitem[Pimentel et~al.(2020{\natexlab{a}})Pimentel, Saphra, Williams, and
  Cotterell]{pimentel2020pareto}
Tiago Pimentel, Naomi Saphra, Adina Williams, and Ryan Cotterell.
\newblock {P}areto probing: {T}rading off accuracy for complexity.
\newblock In Bonnie Webber, Trevor Cohn, Yulan He, and Yang Liu (eds.),
  \emph{Proceedings of the 2020 Conference on Empirical Methods in Natural
  Language Processing (EMNLP)}, pp.\  3138--3153, Online, November
  2020{\natexlab{a}}. Association for Computational Linguistics.
\newblock \doi{10.18653/v1/2020.emnlp-main.254}.
\newblock URL \url{https://aclanthology.org/2020.emnlp-main.254}.

\bibitem[Pimentel et~al.(2020{\natexlab{b}})Pimentel, Valvoda, Maudslay,
  Zmigrod, Williams, and Cotterell]{pimentel2020information}
Tiago Pimentel, Josef Valvoda, Rowan~Hall Maudslay, Ran Zmigrod, Adina
  Williams, and Ryan Cotterell.
\newblock Information-theoretic probing for linguistic structure.
\newblock In Dan Jurafsky, Joyce Chai, Natalie Schluter, and Joel Tetreault
  (eds.), \emph{Proceedings of the 58th Annual Meeting of the Association for
  Computational Linguistics}, pp.\  4609--4622, Online, July
  2020{\natexlab{b}}. Association for Computational Linguistics.
\newblock \doi{10.18653/v1/2020.acl-main.420}.
\newblock URL \url{https://aclanthology.org/2020.acl-main.420}.

\bibitem[Prystawski et~al.(2023)Prystawski, Li, and
  Goodman]{prystawski2023think}
Ben Prystawski, Michael~Y. Li, and Noah~D. Goodman.
\newblock Why think step by step? reasoning emerges from the locality of
  experience, 2023.

\bibitem[Razzhigaev et~al.(2024)Razzhigaev, Mikhalchuk, Goncharova, Oseledets,
  Dimitrov, and Kuznetsov]{razzhigaev2024shape}
Anton Razzhigaev, Matvey Mikhalchuk, Elizaveta Goncharova, Ivan Oseledets,
  Denis Dimitrov, and Andrey Kuznetsov.
\newblock The shape of learning: Anisotropy and intrinsic dimensions in
  transformer-based models, 2024.

\bibitem[Roberts et~al.(2020)Roberts, Raffel, and Shazeer]{roberts2020much}
Adam Roberts, Colin Raffel, and Noam Shazeer.
\newblock How much knowledge can you pack into the parameters of a language
  model?
\newblock In Bonnie Webber, Trevor Cohn, Yulan He, and Yang Liu (eds.),
  \emph{Proceedings of the 2020 Conference on Empirical Methods in Natural
  Language Processing (EMNLP)}, pp.\  5418--5426, Online, November 2020.
  Association for Computational Linguistics.
\newblock \doi{10.18653/v1/2020.emnlp-main.437}.
\newblock URL \url{https://aclanthology.org/2020.emnlp-main.437}.

\bibitem[Rushing \& Nanda(2024)Rushing and Nanda]{rushing2024explorations}
Cody Rushing and Neel Nanda.
\newblock Explorations of self-repair in language models, 2024.

\bibitem[Shin et~al.(2020)Shin, Razeghi, Logan~IV, Wallace, and
  Singh]{shin2020autoprompt}
Taylor Shin, Yasaman Razeghi, Robert~L. Logan~IV, Eric Wallace, and Sameer
  Singh.
\newblock {A}uto{P}rompt: {E}liciting {K}nowledge from {L}anguage {M}odels with
  {A}utomatically {G}enerated {P}rompts.
\newblock In Bonnie Webber, Trevor Cohn, Yulan He, and Yang Liu (eds.),
  \emph{Proceedings of the 2020 Conference on Empirical Methods in Natural
  Language Processing (EMNLP)}, pp.\  4222--4235, Online, November 2020.
  Association for Computational Linguistics.
\newblock \doi{10.18653/v1/2020.emnlp-main.346}.
\newblock URL \url{https://aclanthology.org/2020.emnlp-main.346}.

\bibitem[Tigges et~al.(2023)Tigges, Hollinsworth, Geiger, and
  Nanda]{tigges2023linear}
Curt Tigges, Oskar~John Hollinsworth, Atticus Geiger, and Neel Nanda.
\newblock Linear representations of sentiment in large language models, 2023.

\bibitem[Touvron et~al.(2023)Touvron, Martin, Stone, Albert, Almahairi, Babaei,
  Bashlykov, Batra, Bhargava, Bhosale, Bikel, Blecher, Ferrer, Chen, Cucurull,
  Esiobu, Fernandes, Fu, Fu, Fuller, Gao, Goswami, Goyal, Hartshorn, Hosseini,
  Hou, Inan, Kardas, Kerkez, Khabsa, Kloumann, Korenev, Koura, Lachaux, Lavril,
  Lee, Liskovich, Lu, Mao, Martinet, Mihaylov, Mishra, Molybog, Nie, Poulton,
  Reizenstein, Rungta, Saladi, Schelten, Silva, Smith, Subramanian, Tan, Tang,
  Taylor, Williams, Kuan, Xu, Yan, Zarov, Zhang, Fan, Kambadur, Narang,
  Rodriguez, Stojnic, Edunov, and Scialom]{touvron2023llama}
Hugo Touvron, Louis Martin, Kevin Stone, Peter Albert, Amjad Almahairi, Yasmine
  Babaei, Nikolay Bashlykov, Soumya Batra, Prajjwal Bhargava, Shruti Bhosale,
  Dan Bikel, Lukas Blecher, Cristian~Canton Ferrer, Moya Chen, Guillem
  Cucurull, David Esiobu, Jude Fernandes, Jeremy Fu, Wenyin Fu, Brian Fuller,
  Cynthia Gao, Vedanuj Goswami, Naman Goyal, Anthony Hartshorn, Saghar
  Hosseini, Rui Hou, Hakan Inan, Marcin Kardas, Viktor Kerkez, Madian Khabsa,
  Isabel Kloumann, Artem Korenev, Punit~Singh Koura, Marie-Anne Lachaux,
  Thibaut Lavril, Jenya Lee, Diana Liskovich, Yinghai Lu, Yuning Mao, Xavier
  Martinet, Todor Mihaylov, Pushkar Mishra, Igor Molybog, Yixin Nie, Andrew
  Poulton, Jeremy Reizenstein, Rashi Rungta, Kalyan Saladi, Alan Schelten, Ruan
  Silva, Eric~Michael Smith, Ranjan Subramanian, Xiaoqing~Ellen Tan, Binh Tang,
  Ross Taylor, Adina Williams, Jian~Xiang Kuan, Puxin Xu, Zheng Yan, Iliyan
  Zarov, Yuchen Zhang, Angela Fan, Melanie Kambadur, Sharan Narang, Aurelien
  Rodriguez, Robert Stojnic, Sergey Edunov, and Thomas Scialom.
\newblock Llama 2: Open foundation and fine-tuned chat models, 2023.

\bibitem[Vig et~al.(2020)Vig, Gehrmann, Belinkov, Qian, Nevo, Singer, and
  Shieber]{vig2020investigating}
Jesse Vig, Sebastian Gehrmann, Yonatan Belinkov, Sharon Qian, Daniel Nevo,
  Yaron Singer, and Stuart Shieber.
\newblock Investigating gender bias in language models using causal mediation
  analysis.
\newblock \emph{Advances in neural information processing systems},
  33:\penalty0 12388--12401, 2020.

\bibitem[Virtanen et~al.(2020)Virtanen, Gommers, Oliphant, Haberland, Reddy,
  Cournapeau, Burovski, Peterson, Weckesser, Bright, {van der Walt}, Brett,
  Wilson, Millman, Mayorov, Nelson, Jones, Kern, Larson, Carey, Polat, Feng,
  Moore, {VanderPlas}, Laxalde, Perktold, Cimrman, Henriksen, Quintero, Harris,
  Archibald, Ribeiro, Pedregosa, {van Mulbregt}, and {SciPy 1.0
  Contributors}]{virtanen2020scipy}
Pauli Virtanen, Ralf Gommers, Travis~E. Oliphant, Matt Haberland, Tyler Reddy,
  David Cournapeau, Evgeni Burovski, Pearu Peterson, Warren Weckesser, Jonathan
  Bright, St{\'e}fan~J. {van der Walt}, Matthew Brett, Joshua Wilson, K.~Jarrod
  Millman, Nikolay Mayorov, Andrew R.~J. Nelson, Eric Jones, Robert Kern, Eric
  Larson, C~J Carey, {\.I}lhan Polat, Yu~Feng, Eric~W. Moore, Jake
  {VanderPlas}, Denis Laxalde, Josef Perktold, Robert Cimrman, Ian Henriksen,
  E.~A. Quintero, Charles~R. Harris, Anne~M. Archibald, Ant{\^o}nio~H. Ribeiro,
  Fabian Pedregosa, Paul {van Mulbregt}, and {SciPy 1.0 Contributors}.
\newblock {{SciPy} 1.0: Fundamental Algorithms for Scientific Computing in
  Python}.
\newblock \emph{Nature Methods}, 17:\penalty0 261--272, 2020.
\newblock \doi{10.1038/s41592-019-0686-2}.

\bibitem[Voita \& Titov(2020)Voita and Titov]{voita2020information}
Elena Voita and Ivan Titov.
\newblock Information-theoretic probing with minimum description length.
\newblock In Bonnie Webber, Trevor Cohn, Yulan He, and Yang Liu (eds.),
  \emph{Proceedings of the 2020 Conference on Empirical Methods in Natural
  Language Processing (EMNLP)}, pp.\  183--196, Online, November 2020.
  Association for Computational Linguistics.
\newblock \doi{10.18653/v1/2020.emnlp-main.14}.
\newblock URL \url{https://aclanthology.org/2020.emnlp-main.14}.

\bibitem[Vrande{\v{c}}i{\'c} \& Kr{\"o}tzsch(2014)Vrande{\v{c}}i{\'c} and
  Kr{\"o}tzsch]{vrandevcic2014wikidata}
Denny Vrande{\v{c}}i{\'c} and Markus Kr{\"o}tzsch.
\newblock Wikidata: a free collaborative knowledgebase.
\newblock \emph{Communications of the ACM}, 57\penalty0 (10):\penalty0 78--85,
  2014.

\bibitem[Wang et~al.(2022)Wang, Variengien, Conmy, Shlegeris, and
  Steinhardt]{wang2022interpretability}
Kevin Wang, Alexandre Variengien, Arthur Conmy, Buck Shlegeris, and Jacob
  Steinhardt.
\newblock Interpretability in the wild: a circuit for indirect object
  identification in gpt-2 small, 2022.

\bibitem[Waskom(2021)]{waskom2021seaborn}
Michael~L. Waskom.
\newblock seaborn: statistical data visualization.
\newblock \emph{Journal of Open Source Software}, 6\penalty0 (60):\penalty0
  3021, 2021.
\newblock \doi{10.21105/joss.03021}.
\newblock URL \url{https://doi.org/10.21105/joss.03021}.

\bibitem[Wold(1966)]{wold1966estimation}
Herman Wold.
\newblock Estimation of principal components and related models by iterative
  least squares.
\newblock \emph{Multivariate analysis}, pp.\  391--420, 1966.

\bibitem[Wold et~al.(2001)Wold, Sj{\"o}str{\"o}m, and Eriksson]{wold2001pls}
Svante Wold, Michael Sj{\"o}str{\"o}m, and Lennart Eriksson.
\newblock Pls-regression: a basic tool of chemometrics.
\newblock \emph{Chemometrics and intelligent laboratory systems}, 58\penalty0
  (2):\penalty0 109--130, 2001.

\bibitem[Wolf et~al.(2020)Wolf, Debut, Sanh, Chaumond, Delangue, Moi, Cistac,
  Rault, Louf, Funtowicz, Davison, Shleifer, von Platen, Ma, Jernite, Plu, Xu,
  Le~Scao, Gugger, Drame, Lhoest, and Rush]{wolf2020transformers}
Thomas Wolf, Lysandre Debut, Victor Sanh, Julien Chaumond, Clement Delangue,
  Anthony Moi, Pierric Cistac, Tim Rault, Remi Louf, Morgan Funtowicz, Joe
  Davison, Sam Shleifer, Patrick von Platen, Clara Ma, Yacine Jernite, Julien
  Plu, Canwen Xu, Teven Le~Scao, Sylvain Gugger, Mariama Drame, Quentin Lhoest,
  and Alexander Rush.
\newblock Transformers: State-of-the-art natural language processing.
\newblock In Qun Liu and David Schlangen (eds.), \emph{Proceedings of the 2020
  Conference on Empirical Methods in Natural Language Processing: System
  Demonstrations}, pp.\  38--45, Online, October 2020. Association for
  Computational Linguistics.
\newblock \doi{10.18653/v1/2020.emnlp-demos.6}.
\newblock URL \url{https://aclanthology.org/2020.emnlp-demos.6}.

\bibitem[Youssef et~al.(2023)Youssef, Kora{\c{s}}, Li, Schl{\"o}tterer, and
  Seifert]{youssef2023facts}
Paul Youssef, Osman Kora{\c{s}}, Meijie Li, J{\"o}rg Schl{\"o}tterer, and
  Christin Seifert.
\newblock Give me the facts! a survey on factual knowledge probing in
  pre-trained language models.
\newblock In Houda Bouamor, Juan Pino, and Kalika Bali (eds.), \emph{Findings
  of the Association for Computational Linguistics: EMNLP 2023}, pp.\
  15588--15605, Singapore, December 2023. Association for Computational
  Linguistics.
\newblock \doi{10.18653/v1/2023.findings-emnlp.1043}.
\newblock URL \url{https://aclanthology.org/2023.findings-emnlp.1043}.

\bibitem[Zhang \& Nanda(2024)Zhang and Nanda]{zhang2024best}
Fred Zhang and Neel Nanda.
\newblock Towards best practices of activation patching in language models:
  Metrics and methods, 2024.

\bibitem[Zhong et~al.(2021)Zhong, Friedman, and Chen]{zhong2021factual}
Zexuan Zhong, Dan Friedman, and Danqi Chen.
\newblock Factual probing is [{MASK}]: Learning vs. learning to recall.
\newblock In Kristina Toutanova, Anna Rumshisky, Luke Zettlemoyer, Dilek
  Hakkani-Tur, Iz~Beltagy, Steven Bethard, Ryan Cotterell, Tanmoy Chakraborty,
  and Yichao Zhou (eds.), \emph{Proceedings of the 2021 Conference of the North
  American Chapter of the Association for Computational Linguistics: Human
  Language Technologies}, pp.\  5017--5033, Online, June 2021. Association for
  Computational Linguistics.
\newblock \doi{10.18653/v1/2021.naacl-main.398}.
\newblock URL \url{https://aclanthology.org/2021.naacl-main.398}.

\end{thebibliography}
\bibliographystyle{colm2024_conference}

\appendix
\newpage

\section{Data sample}
\label{sec:data_sample}

\begin{appendixtbl}
	\centering
	\adjustbox{max width=\linewidth}{
		\begin{tabular}{llp{11em}lp{20em}lll}
\toprule
Property & Prop. ID & Entity & Entity ID & Prompt & Value & Unit\\
\midrule
birthyear & P569 & Nina Foch & Q235632 & In what year was Nina Foch born? & 1924 & annum\\
birthyear & P569 & Geoffrey Holder & Q945691 & In what year was Geoffrey Holder born? & 1930 & annum\\
birthyear & P569 & Harriette L. Chandler & Q5664432 & In what year was Harriette L. Chandler born? & 1937 & annum\\
birthyear & P569 & Gabriel García Márquez & Q5878 & In what year was Gabriel García Márquez born? & 1927 & annum\\
birthyear & P569 & Norman Schwarzkopf Jr. & Q310188 & In what year was Norman Schwarzkopf Jr. born? & 1934 & annum\\
birthyear & P569 & Paul de Vos & Q2610964 & In what year was Paul de Vos born? & 1590 & annum\\
birthyear & P569 & Nicolas Carnot & Q181685 & In what year was Nicolas Carnot born? & 1796 & annum\\
birthyear & P569 & Steve Harvey & Q2347009 & In what year was Steve Harvey born? & 1957 & annum\\
birthyear & P569 & Tommy Lawton & Q726272 & In what year was Tommy Lawton born? & 1919 & annum\\
birthyear & P569 & Hans von Bülow & Q155540 & In what year was Hans von Bülow born? & 1830 & annum\\
\midrule
death year & P570 & Johannes R. Becher & Q58057 & In what year did Johannes R. Becher die? & 1958 & annum\\
death year & P570 & Friedrich Georg Wilhelm von Struve & Q57164 & In what year did Friedrich Georg Wilhelm von Struve die? & 1864 & annum\\
death year & P570 & Pierre Boulez & Q156193 & In what year did Pierre Boulez die? & 2016 & annum\\
death year & P570 & Giovanni da Palestrina & Q179277 & In what year did Giovanni da Palestrina die? & 1594 & annum\\
death year & P570 & Abdurrauf Fitrat & Q317907 & In what year did Abdurrauf Fitrat die? & 1938 & annum\\
death year & P570 & Lucian Freud & Q154594 & In what year did Lucian Freud die? & 2011 & annum\\
death year & P570 & Akseli Gallen-Kallela & Q170068 & In what year did Akseli Gallen-Kallela die? & 1931 & annum\\
death year & P570 & Spock & Q16341 & In what year did Spock die? & 2263 & annum\\
death year & P570 & William Orpen & Q922483 & In what year did William Orpen die? & 1931 & annum\\
death year & P570 & Carlos Santiago Mérida & Q1043100 & In what year did Carlos Santiago Mérida die? & 1984 & annum\\
\midrule
population & P1082 & Akhisar & Q209905 & What is the population of Akhisar? & 173026 & 1\\
population & P1082 & Tripura & Q1363 & What is the population of Tripura? & 3665958 & 1\\
population & P1082 & Albert & Q30940 & What is the population of Albert? & 9930 & 1\\
population & P1082 & High Wycombe & Q64116 & What is the population of High Wycombe? & 120256 & 1\\
population & P1082 & Plön & Q497060 & What is the population of Plön? & 8914 & 1\\
population & P1082 & Republika Srpska & Q11196 & What is the population of Republika Srpska? & 1228423 & 1\\
population & P1082 & Lebanese & Q2606511 & What is the population of Lebanese? & 8000000 & 1\\
population & P1082 & Geraardsbergen & Q499532 & What is the population of Geraardsbergen? & 33403 & 1\\
population & P1082 & Gorzów Wielkopolski & Q104731 & What is the population of Gorzów Wielkopolski? & 124295 & 1\\
population & P1082 & Harran & Q199547 & What is the population of Harran? & 47606 & 1\\
\midrule
evelation & P2044 & Sondrio & Q6274 & How high is Sondrio? & 360 & metre\\
evelation & P2044 & Rio Branco & Q171612 & How high is Rio Branco? & 158 & metre\\
evelation & P2044 & Demmin & Q50960 & How high is Demmin? & 8 & metre\\
evelation & P2044 & Cetinje & Q173338 & How high is Cetinje? & 650 & metre\\
evelation & P2044 & Highland Park & Q576671 & How high is Highland Park? & 503 & metre\\
evelation & P2044 & Gozo & Q170488 & How high is Gozo? & 195 & metre\\
evelation & P2044 & Saint-Jean-de-Maurienne & Q208860 & How high is Saint-Jean-de-Maurienne? & 566 & metre\\
evelation & P2044 & Butte & Q467664 & How high is Butte? & 1688 & metre\\
evelation & P2044 & Cottbus & Q3214 & How high is Cottbus? & 76 & metre\\
evelation & P2044 & Mahilioŭ Region & Q189822 & How high is Mahilioŭ Region? & 191 & metre\\
\midrule
longitude & P625.long & Korean Empire & Q28233 & What is the longitude of Korean Empire? & 126.98 & degree\\
longitude & P625.long & Pine Bluff & Q80012 & What is the longitude of Pine Bluff? & -92.00 & degree\\
longitude & P625.long & Tegernsee & Q260130 & What is the longitude of Tegernsee? & 11.76 & degree\\
longitude & P625.long & Old Cölln & Q269622 & What is the longitude of Old Cölln? & 13.40 & degree\\
longitude & P625.long & Cambridge & Q49111 & What is the longitude of Cambridge? & -71.11 & degree\\
longitude & P625.long & Stryn & Q5223 & What is the longitude of Stryn? & 6.86 & degree\\
longitude & P625.long & Ciudad Real Province & Q54932 & What is the longitude of Ciudad Real Province? & -4.00 & degree\\
longitude & P625.long & Santa Catarina & Q41115 & What is the longitude of Santa Catarina? & -50.49 & degree\\
longitude & P625.long & Wake Forest University & Q392667 & What is the longitude of Wake Forest University? & -80.28 & degree\\
longitude & P625.long & West Lothian & Q204940 & What is the longitude of West Lothian? & -3.50 & degree\\
\midrule
latitude & P625.lat & Küsnacht & Q69216 & What is the latitude of Küsnacht? & 47.32 & degree\\
latitude & P625.lat & Mount Jerome Cemetery & Q917854 & What is the latitude of Mount Jerome Cemetery? & 53.32 & degree\\
latitude & P625.lat & Dayton Children's Hospital & Q5243510 & What is the latitude of Dayton Children's Hospital? & 39.77 & degree\\
latitude & P625.lat & Le Flore County & Q495944 & What is the latitude of Le Flore County? & 34.90 & degree\\
latitude & P625.lat & Czechoslovakia & Q33946 & What is the latitude of Czechoslovakia? & 50.08 & degree\\
latitude & P625.lat & Pembroke College & Q956501 & What is the latitude of Pembroke College? & 52.20 & degree\\
latitude & P625.lat & Hayward & Q491114 & What is the latitude of Hayward? & 37.67 & degree\\
latitude & P625.lat & Banaskantha district & Q806125 & What is the latitude of Banaskantha district? & 24.17 & degree\\
latitude & P625.lat & Corbeil-Essonnes & Q208812 & What is the latitude of Corbeil-Essonnes? & 48.61 & degree\\
latitude & P625.lat & Elbasan & Q114257 & What is the latitude of Elbasan? & 41.11 & degree\\
\bottomrule
\end{tabular}

	}
	\caption{%
Random sample of the entities used in our experiments, along with corresponding numeric attributes and prompts.
	Entities, their English labels, and numeric attributes for each property are extracted from an April 2022 dump of Wikidata (\texttt{wikidata-20220421-all}).
	In many cases Wikidata contains multiple values for a given numeric attribute, e.g., reflecting chronological change such as the population of a city, or owing to conflicting sources.
	In such cases we take the mode of the distribution as gold value.
	We also filter out quantities with non-standard units, such as elevations measured in feet.%
	}
	\label{tbl:data_sample_10}
\end{appendixtbl}

\section{Regression on entity representations: Additional figures}
\label{sec:more_regression_results}

\newcommand\regrappfig[2]{
\begin{appendixfig}
	\plssubfigs{#1}
	\caption{Regression curves for #2. See explanation in Figure~\ref{fig:pls_results}.}
\end{appendixfig}
}

\regrappfig{Llama27bhf}{\llamasmall}

\regrappfig{falcon7binstruct}{\falconsmall}

\regrappfig{Mistral7B}{\mistralsmall}

\section{Regression on entity representations: Additional analysis}
\label{sec:dim_reduction_data}

\begin{appendixtbl}
	\centering
	\adjustbox{max width=\linewidth}{
		\begin{tabular}{lllrrrrrrr}
\toprule
Property & Model & $R^2$ & $C\left[max R^2\right]$ & $C\left[\ge 0.95 R^2\right]$ & $C\left[\ge 0.90 R^2\right]$ & $C\left[\ge 0.80 R^2\right]$ & $C\left[\ge 0.70 R^2\right]$ & $C\left[\ge 0.60 R^2\right]$ & $C\left[\ge 0.50 R^2\right]$ \\
\midrule
birthyear (P569) & Falcon 7B & 0.75 & 4 & 2 & 2 & 2 & 1 & 1 & 1\\
birthyear (P569) & Llama 2 13B & 0.91 & 7 & 4 & 3 & 2 & 2 & 2 & 1\\
birthyear (P569) & Llama 2 7B & 0.90 & 11 & 6 & 4 & 3 & 2 & 2 & 1\\
birthyear (P569) & Mistral 7B & 0.89 & 4 & 3 & 2 & 2 & 2 & 1 & 1\\
death year (P570) & Falcon 7B & 0.61 & 2 & 2 & 2 & 2 & 1 & 1 & 1\\
death year (P570) & Llama 2 13B & 0.84 & 12 & 4 & 3 & 2 & 2 & 1 & 1\\
death year (P570) & Llama 2 7B & 0.82 & 11 & 4 & 4 & 3 & 2 & 1 & 1\\
death year (P570) & Mistral 7B & 0.80 & 4 & 3 & 3 & 2 & 2 & 1 & 1\\
latitude (P625.lat) & Falcon 7B & 0.67 & 6 & 3 & 3 & 3 & 2 & 2 & 2\\
latitude (P625.lat) & Llama 2 13B & 0.82 & 10 & 5 & 4 & 3 & 3 & 2 & 2\\
latitude (P625.lat) & Llama 2 7B & 0.83 & 10 & 5 & 3 & 2 & 2 & 2 & 2\\
latitude (P625.lat) & Mistral 7B & 0.79 & 9 & 4 & 3 & 3 & 2 & 2 & 2\\
longitude (P625.long) & Falcon 7B & 0.74 & 7 & 5 & 3 & 3 & 2 & 2 & 2\\
longitude (P625.long) & Llama 2 13B & 0.79 & 17 & 6 & 5 & 3 & 3 & 2 & 2\\
longitude (P625.long) & Llama 2 7B & 0.83 & 9 & 5 & 3 & 3 & 2 & 2 & 2\\
longitude (P625.long) & Mistral 7B & 0.78 & 6 & 5 & 3 & 3 & 2 & 2 & 1\\
population (P1082) & Falcon 7B & 0.67 & 4 & 3 & 3 & 2 & 1 & 1 & 1\\
population (P1082) & Llama 2 13B & 0.79 & 5 & 4 & 4 & 2 & 2 & 1 & 1\\
population (P1082) & Llama 2 7B & 0.73 & 5 & 4 & 3 & 2 & 2 & 1 & 1\\
population (P1082) & Mistral 7B & 0.76 & 5 & 4 & 2 & 2 & 1 & 1 & 1\\
evelation (P2044) & Falcon 7B & 0.23 & 2 & 2 & 2 & 2 & 2 & 1 & 1\\
evelation (P2044) & Llama 2 13B & 0.43 & 3 & 2 & 2 & 2 & 2 & 2 & 1\\
evelation (P2044) & Llama 2 7B & 0.37 & 2 & 2 & 2 & 2 & 2 & 1 & 1\\
evelation (P2044) & Mistral 7B & 0.41 & 3 & 3 & 2 & 2 & 2 & 1 & 1\\
\bottomrule
\end{tabular}

	}
	\caption{Number of partial least squares regression components $C\left[T\right]$ required for a given goodness of fit $T$, found using the experimental setup described in \secref{sec:correlation}.
	For example, the $C\left[\ge0.95R^2\right]$ column shows the number of components required to reach 95 percent of the maximum goodness of fit for the respective property and model.
	From this column we can read that, e.g., two components of Falcon 7B's activation space are sufficient to reach 95 percent of the maximum goodness of fit when predicting the birthyear of entities, indicating that this property is almost entirely encoded in a two-dimensional subspace of this model's activation space.
	}
	\label{tbl:dim_reduction_n_comp}
\end{appendixtbl}

\section{PLS projections of entity representations: Additional figures}
\label{sec:more_dimred_results}

\newcommand\dimredappfig[2]{
\begin{appendixfig}
	\dimredsubfigs{#1}
	\caption{PLS projections of #2 entity representations. See explanation in Figure~\ref{fig:dim_red}.}
\end{appendixfig}
}

\dimredappfig{Llama27bhf}{\llamasmall}

\dimredappfig{falcon7binstruct}{\falconsmall}

\dimredappfig{Mistral7B}{\mistralsmall}

\section{Choice of probing and edit locus}
\label{sec:edit_locus}

\begin{appendixfig}
	\includegraphics[width=0.6\textwidth]{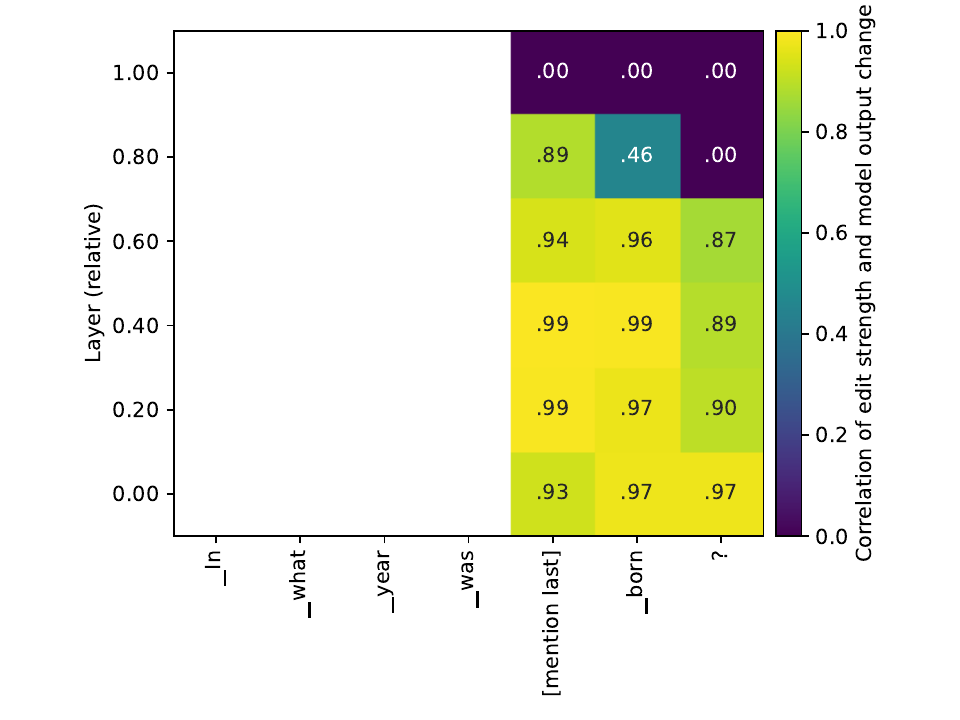}
	\caption{%
Results of a cursory search for the best probing and edit locus, using \llamasmall.
Varying token position and layer, we edit the hidden state at this locus as described in \secref{sec:causation} and record the Spearman correlation between edit strength and the change in the quantity (here: birthyear) expressed by the model.
Correlation is highest ($0.99$) in the region between layers $0.2$ and $0.4$ and the last subword token of the entity mention and the following token.
Based on this, we choose the last mention token and the middle point at layer $l = 0.3$ as locus for the regression experiments in \secref{sec:correlation} and activation patching experiments in \secref{sec:causation}, across all numeric properties and LMs, but acknowledge that a more exhaustive search would likely find better probing and edit loci.%
	\label{fig:edit_locus}
	}
\end{appendixfig}

\section{Edit curves for additional language models}
\label{sec:more_edit_curves}

\begin{appendixfig}
	\editsubfigs{llama27b}
	\caption{Effect of activation patching along property-specific directions across several numeric properties with \llamasmall. See explanation in Figure~\ref{fig:effects}.
	}
	\label{fig:effects_llama_2_7b}
\end{appendixfig}

\begin{appendixfig}
	\editsubfigs{falcon7b}
	\caption{Effect of activation patching along property-specific directions across several numeric properties with \falconsmall \citep{almazrouei2023falcon}. See explanation in Figure~\ref{fig:effects}.
	}
	\label{fig:effects_falcon_7b}
\end{appendixfig}

\begin{appendixfig}
	\editsubfigs{mistral7b}
	\caption{Effect of activation patching along property-specific directions across several numeric properties with \model{Mistral 7B} \citep{jiang2023mistral}. See explanation in Figure~\ref{fig:effects}.
	}
	\label{fig:effects_mistral_7b}
\end{appendixfig}

\section{Software}

The following is a list of the main libraries used in this work:

\begin{itemize}
	\item Numpy \citep{harris2020array}
	\item Scikit-learn \citep{pedregosa2011scikit}
	\item Pytorch \citep{paszke2019pytorch}
	\item Transformers \citep{wolf2020transformers}
	\item seaborn \citep{waskom2021seaborn}
	\item Matplotlib \citep{hunter2007matplotlib}
	\item SciPy \citep{virtanen2020scipy}
	\item Pandas \citep{reback2020pandas}
\end{itemize}

We thank all authors and the open source community in general for creating and maintaining publicly and freely available software.

\end{document}